\documentclass{article}

\PassOptionsToPackage{sort&compress,numbers,super}{natbib}

\usepackage[preprint]{neurips_2026}

\usepackage[utf8]{inputenc} 
\usepackage[T1]{fontenc}    
\usepackage{hyperref}       
\usepackage{url}            
\usepackage{booktabs}       
\usepackage{amsfonts}       
\usepackage{nicefrac}       
\usepackage{microtype}      
\usepackage{xcolor}
\usepackage{subcaption}
\usepackage{longtable}
\usepackage{array}
\usepackage{xcolor}
\usepackage{graphicx}%
\usepackage{multirow}%
\usepackage{amsmath,amssymb,amsfonts}%
\usepackage{amsthm}%
\usepackage{mathrsfs}%
\usepackage[title]{appendix}%
\usepackage{textcomp}%
\usepackage{booktabs}%
\usepackage{algorithm}%
\usepackage{algorithmicx}%
\usepackage{algpseudocode}%
\usepackage{listings}%
\usepackage[numbers,sort&compress]{natbib}
\usepackage{chemformula} 

\theoremstyle{plain}%
\theoremstyle{remark}%

\theoremstyle{definition}%

\title{Agentic generation of verifiable rules for deterministic, self-expanding reaction classification}

\author{
  Daniel Armstrong\textsuperscript{1}, Maarten Dobbelaere\textsuperscript{1,2},\\
  \textbf{Valentas Olikauskas\textsuperscript{1,3}, Helena Avila\textsuperscript{1,3}, Octavian Susanu\textsuperscript{1},}\\
  \textbf{J\'{e}r\^{o}me Waser\textsuperscript{1,3}, Philippe Schwaller\textsuperscript{1,3}} \\
  \textsuperscript{1}\'Ecole Polytechnique F\'{e}d\'{e}rale de Lausanne (EPFL), Switzerland \\
  \textsuperscript{2}Ghent University, Belgium
\\
  \textsuperscript{3}National Centre of Competence in Research (NCCR) Catalysis, Switzerland \\
  \texttt{\{daniel.armstrong,philippe.schwaller\}@epfl.ch} \\
}

\begin{document}

\maketitle

\begin{abstract}
Computer-assisted synthesis planning breaks target molecules into accessible precursors using large libraries of reaction rules that assign each transformation a deterministic, interpretable label. But chemistry is long-tailed, making manual encoding intractable, and existing tools rely on fixed rulesets that cannot adapt to new chemistries. Here we present a fully automated pipeline in which a multi-agent framework of large language models (LLMs) classifies reactions and writes the rules themselves across 665,901 US patent reactions, generating each rule under a verification loop that tests it against the corpus. It expands a standard taxonomy from  68 to 14,073 classes without human curation. With a lightweight fingerprint classifier, it classifies 97.7\% of unseen reactions, matching a leading proprietary classifier while resolving chemistry more finely and extending on demand to chemistry outside its training distribution. The result is a living reactivity database and a general route to turning generative models into reliable, self-expanding symbolic systems.
\end{abstract}

\section{Introduction}\label{sec1}

Chemical synthesis planning, in which a complex target molecule is decomposed recursively into simpler building blocks, remains a core challenge in drug discovery and materials design. In their seminal 1969 work, Corey and Wipke proposed computer-assisted synthesis planning (CASP), applying encoded mechanistic rules to suggest synthetic pathways \citep{corey1967general, corey1969computer, corey1972computer, corey1985computer}. To date, this rule-based approach remains common; the most comprehensive implementations contain tens of thousands of manually designed reaction rules with hardcoded protection and incompatibility logic \citep{grzybowski2018chematica}. Yet the distribution of chemical reactions follows a power law \citep{zipf2013psycho, szymkuc2016computer}, and the long tail of rare transformations, each demanding the same careful encoding as a commonplace amide coupling, makes exhaustive manual coverage practically intractable. Automating the extraction and generalisation of reaction rules is particularly timely, as the utilisation of Large Language Models (LLMs) in chemistry matures \citep{bran2023chemcrow, jablonka2023gpt}. Such models may offer a new way to encode the synthetic toolbox, replacing human logic with the automatic inference of symbolic reaction transforms from chemical data.

The automation of reaction rule generation requires solving two distinct problems. First, a reaction must be assigned to a named class within a structured taxonomy. Carey et al. \citep{carey2006analysis} introduced a semantic hierarchy of ten superclasses to analyse common industrial pharmaceutical transformations, and together with the subsequent medicinal chemistry analysis by Roughley and Jordan \citep{roughley2011medicinal}, this work informed the Royal Society of Chemistry's RXNO ontology of named reactions \cite{rxno}. Second, the generalised transformation within each class must be encoded as a computable reaction rule. Such rules are typically expressed as SMIRKS, a text based format which encodes a graph transformation around the reaction centre and its local atomic environment \cite{daylightsmirks}. However, constructing these patterns depends heavily on chemical intuition; even a single transformation requires careful specification of atom mappings, stereochemical constraints, and functional group compatibility, and the effort scales poorly to the thousands of reaction types observed in practice.

Data-driven approaches to reaction classification have emerged over the past decade, enabled in large part by the publicly available USPTO reaction dataset introduced by \cite{Lowe2012, Lowe2017}. This corpus, which has also supported work on outcome prediction \cite{Jin2017predicting,coley2017prediction,Schwaller2018found,schwaller2019molecular} and multistep route planning \cite{coley2017computer,Segler2017neural,segler2018planning,schwaller2020predicting}, provides the scale of labelled data needed to train and evaluate classifiers \citep{probst2021molecular,Schwaller2021mapping}; yet these methodologies still depend on a proprietary tool to determine their ground-truth labels. This dependence exposes a deeper limitation, since existing classification tools, whether the \textit{de facto} standard NameRXN \citep{schneider2016big,Jin2017predicting,joung2025electron} or open-source alternatives such as Rxn-INSIGHT \citep{rxn-insight}, fail on both fronts: they rely on fixed ontologies that cannot accommodate transformations absent from their original design, and on manually encoded rule sets with inherently limited template coverage. This sparsity directly impairs generative molecular design; because synthesisability-constrained generation relies on template libraries to steer models, tools tethered to static rule sets restrict exploration to historically established chemical space. A dynamically expandable, granular reaction naming and encoding system could address this bottleneck. By generating rules for novel chemistry on demand, researchers could explicitly condition generative models to sample and map the previously inaccessible regions of chemical space unlocked by newly discovered transformations.

The second problem, scalable rule encoding, remains unsolved. Template-free methods, which employ graph- or language-based models to predict reactants directly from a target product, have shown strong performance for reaction outcome prediction \cite{Jin2017predicting,Schwaller2018found,schwaller2019molecular} and retrosynthetic planning \cite{coley2017computer,Segler2017neural,segler2018planning,schwaller2020predicting}, bypassing template creation entirely \citep{schwaller2022machine}. However, these approaches suffer from generating physically invalid molecules and transformations \citep{gil2023holistic}, a weakness that is particularly consequential for reaction classification. Here, rule-based matching retains a distinctive advantage: a SMIRKS pattern either matches a reaction or it does not, so a successful match provides an unambiguous, deterministic assignment. Neural reaction classifiers, by contrast, return probability distributions over classes, and even highly accurate models will occasionally mis-assign a reaction, with such errors propagating silently into downstream tasks. This makes rule-based matching especially attractive for classification, provided the rule library can be made sufficiently comprehensive; yet no automated method has yet produced a validated library at the scale required.

A further challenge lies in the redundancy of reaction templates: a single transformation such as amide bond formation may be captured by hundreds of distinct templates in a corpus, each with subtle structural variations, yet only a handful are needed to minimally describe the chemistry. LLMs offer a natural mechanism to automate this rule distillation process~\citep{vaswani2017attention, radford2019language, openai_gpt4, google_gemini}. While these models demonstrate strong knowledge of chemical synthesis, including structure, reaction feasibility, and strategic planning~\citep{bran2025chemical, xuan2025synthelite, armstrong2025synthstrategy, hassen2025atom}, they have shown notable difficulty with the related SMILES notation~\citep{bran2025chemical}, and the generation of valid reaction SMARTS has not been demonstrated. Furthermore, it has not been proven whether this knowledge can be applied with sufficient precision and consistency across hundreds of thousands of reactions. Even where such precision is achievable, the per-reaction inference cost of current LLMs makes real-time classification impractical, motivating hybrid approaches that combine LLM-distilled knowledge with fast neuro-symbolic matching.

Here we show that LLMs can perform detailed reaction classification and reaction SMARTS generation across a large section of the USPTO corpus ($665{,}901$ reactions) with near-human label accuracy. Starting from the RXNO ontology as a seed, we employ an LLM to iteratively classify reactions while dynamically expanding the taxonomy to capture chemistry not represented in the original hierarchy, growing from $68$ seed classes to $14{,}073$ through successive refinement. For each class, the LLM generalises reaction SMIRKS at intermediate hierarchical tiers; these are validated against the corpus through an autonomous refinement loop that eliminates false positives without sacrificing recall. The resulting SMIRKS, combined with a lightweight MLP and template matching, enable a fast, deterministic assignment of reactions to their taxonomical position at $97.7\%$ strict-match accuracy at the third hierarchical level. This two-layer architecture addresses the competing demands of coverage and speed: the deterministic SMIRKS layer provides rapid classification suitable for integration into real-time synthesis planning and molecular design workflows, while the LLM layer can be invoked on demand to classify reactions that fall outside existing template coverage, automatically proposing new taxonomy entries as needed.

\begin{figure}[htbp]
    \centering
    \includegraphics[width=1.0\textwidth]{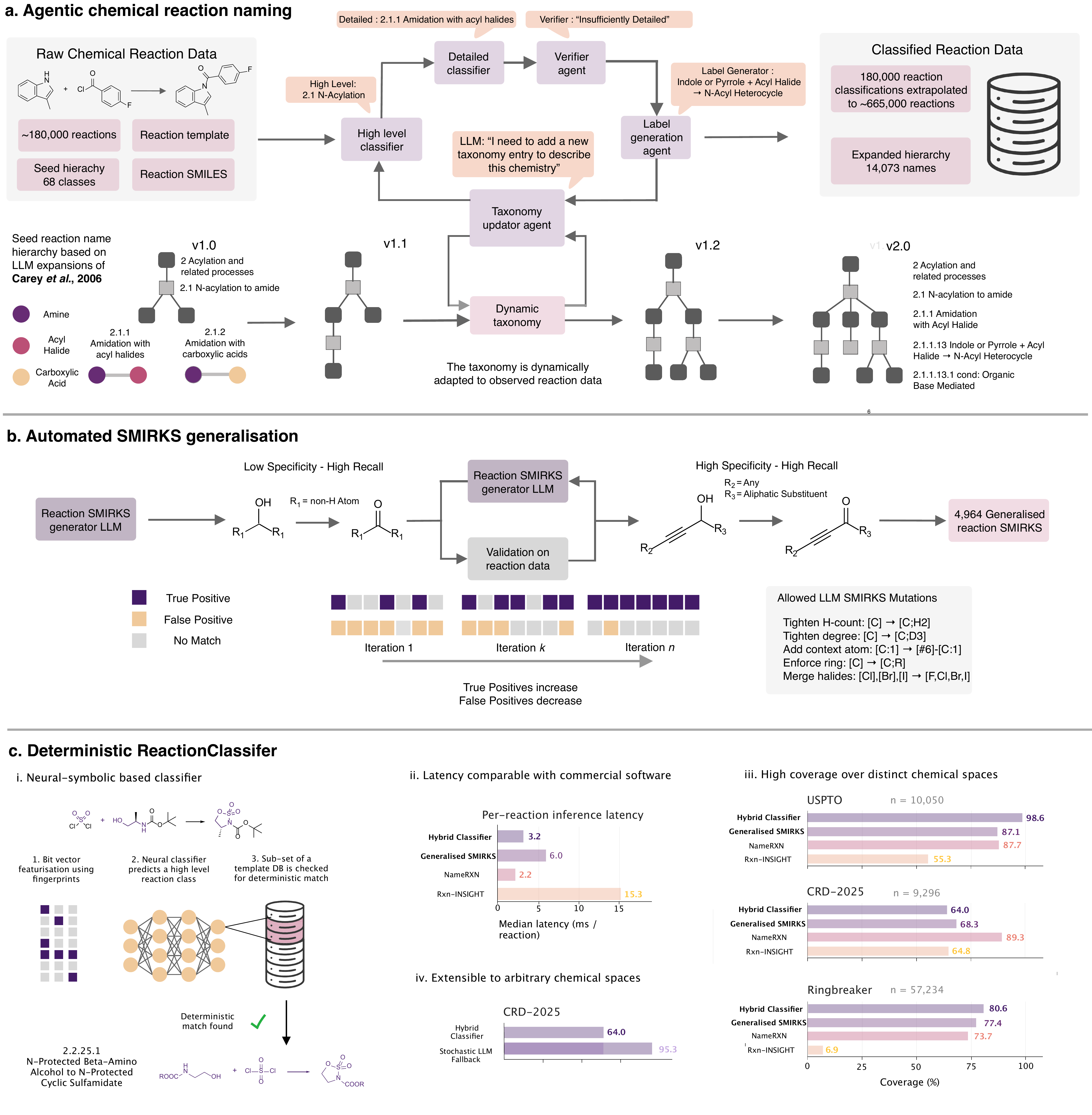}
    \caption{a. The multi-agent LLM framework for dynamic taxonomy expansion. Raw reaction data and a seed hierarchy are evaluated by sequential high-level and detailed classification agents. A verifier agent audits the proposed labels; when existing categories are insufficient or imprecise, a label generation agent proposes new classifications. A taxonomy updater integrates these proposals, dynamically expanding the hierarchy to capture unrepresented chemistry without human intervention, resulting in an expanded reaction taxonomy. b. The iterative refinement loop for reaction template generation. An LLM drafts initial, broad SMIRKS patterns (low specificity, high recall) which undergo automated validation against reaction data. Through successive iterations, the LLM applies chemically grounded structural mutations to minimize false positives while maximizing true positive recall, yielding a robust database of generalized, high-specificity reaction templates. c. A highlight of ReactionClassifiers neuro-symbolic architecture, latency and coverage over diverse chemical spaces.}
    \label{fig:classification_main}
\end{figure}

\section{Results and Discussion}\label{sec2}

\subsection{Large language model analysis of chemical data}

The core challenge in applying LLMs to large-scale reaction annotation is that a single model call cannot reliably classify hundreds of thousands of reactions in one pass. Even state-of-the-art long-context models exhibit systematic degradation when retrieving and reasoning over information positioned deep within their input window where accuracy follows a U-shaped curve, peaking for content at the beginning and end of the context while dropping sharply in the middle \citep{liu2024lost}. Subsequent work has shown that this performance loss is not solely positional but scales with input length itself, persisting even when the relevant evidence is placed immediately before the query \citep{du2025context}. For a corpus of millions of reactions that encompass thousands of mechanistically distinct transformation types, it is therefore expected that naïvely embedding the entire taxonomy and reaction set into a single prompt produces unreliable assignments.

To minimise such errors, we decompose the classification task into five specialised LLM agents, with the overall architecture shown in Figure \ref{fig:classification_main} a. This design leverages the Chain-of-Verification paradigm \citep{dhuliawala2024chain}, drawing on the principle that isolating initial generation from independent verification significantly reduces factual hallucination. To ensure mechanistically equivalent transformations yield identical labels, reactions are first grouped into template-level cohorts based on shared retrosynthetic templates, as extracted by RDChiral in AiZynthTrain \citep{3aizynthtrain}. A coarse-grained hierarchy agent then maps each cohort to one of 68 core sub-classes, which a detailed agent refines to the finest applicable level of detail in the full taxonomy. If the cohort cannot be assigned to an entry in the taxonomy, it is flagged as such. A separate verification agent then checks these assignments against the reaction SMILES and template. The verifier either accepts the detailed classification, or rejects it and passes it on to the next stage.

Following verification, processing a reaction corpus requires a dynamic taxonomy. Fixed classification schemes, such as the RXNO ontology or NameRXN, inevitably encounter novel chemistry that forces uninformative "unclassified" tags or outright mis-assignments. We overcome this by allowing the taxonomy to expand adaptively based on emergent data patterns rather than relying solely on the model's prior knowledge \citep{kargupta2025taxoadapt}. To achieve this, we developed a generator-aggregator architecture: when a reaction is failed by the verifier, a generator agent proposes a new taxonomy entry complete with a code, name, and hierarchical position. This cannot modify the existing taxonomy, only add to it. An aggregator agent then deduplicates these proposals, resolves structural conflicts, and atomically updates the living hierarchy. This automated loop allowed our system to expand organically from the initial 68 sub classes defined by \citep{carey2006analysis} to 14,073 distinct class labels.

The pipeline classified 179,495 reactions across 42,125 templates. Assuming that all reactions sharing a template inherit the same taxonomic classification, we extrapolate via template identity to yield 665,901 labeled reactions in the full dataset. The result is a hierarchical classification with granularity greater than manual expert annotation, but at a scale and speed that would be infeasible by hand.

To validate the LLM labelling, we aimed to gain an estimate of the labelling noise compared to a human expert baseline, NameRXN. As finer taxonomies may incur higher per-label noise, before evaluating classification accuracy, we first controlled for this granularity effect by truncating the LLM-generated hierarchy to match the breadth of NameRXN (version 3.7.0) over an identical reaction set. This truncation yielded 823 classes for the LLM pipeline versus 1,029 for NameRXN, enabling a direct comparison of label reliability without requiring complex semantic alignment between the two distinct systems.With the taxonomies aligned, we assess the residual annotation error using the confident learning framework \citep{northcutt2021confident}. A lightweight neural classifier, trained on reaction fingerprints via five-fold stratified cross-validation, generated out-of-fold class probabilities. Because each reaction was scored by a model instance that had never seen its original label, the classifier could flag anomalous examples based purely on structural patterns without memorising the pipeline's assignments. This self-confidence procedure estimated a label noise rate of 2.19\% for the LLM taxonomy (97.81\% accuracy), compared with 0.59\% for NameRXN (99.41\%).However, statistical confidence limit often confuse genuine misclassification with ambiguity at taxonomy boundaries.

\begin{table}[ht]
\centering
\caption{Estimated label noise from confident learning. Noise rate denotes the fraction of training examples flagged as likely mislabelled; accuracy is the complementary.}
\label{tab:label_noise}
\begin{tabular}{lccc}
\toprule
Label source & Classes & Noise (\%) & Accuracy (\%) \\
\midrule
LLM (level 3) & 823  & 2.19 & 97.81 \\
NameRXN        & 1029 & 0.59 & 99.41 \\
\bottomrule
\end{tabular}
\end{table}
\begin{figure}[htb]
    \centering
    \includegraphics[width=0.95\textwidth]{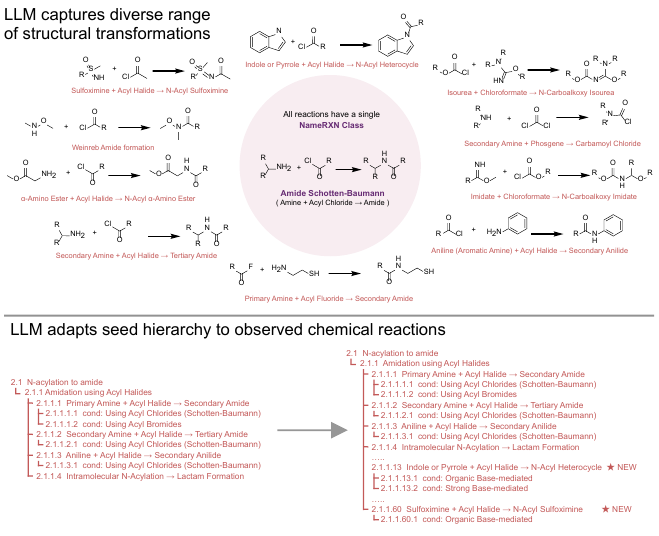}
    \caption{In \textbf{a.} we show an example of a set of reactions classified into a single bucket by NameRXN, alongside distinct classes proposed by our LLM based methodology. In part \textbf{b.} we show how the taxonomy is adapted by the LLM to observed chemical data.}
    \label{fig:namerxn_llm_comparision}
\end{figure}
To determine how much of this apparent noise reflects genuine errors versus taxonomy-boundary ambiguity, we focused on the LLM-labelled test set (n = 30,802) and applied an LLM-as-judge framework. A label-prediction conflict arises when the MLP classifier's predicted class diverges from the original label assigned by the LLM pipeline. Of the 600 reactions flagged by confident learning (1.95\%), 321 exhibited such conflicts and were used for adjudication. For each pair, the judge model was provided with the two class definitions and asked to assess whether the classes share genuine mechanistic or substrate overlap. To mitigate the circularity of using an LLM to adjudicate labels originally generated by the Gemini-based pipeline, the judge operated without access to the original classifier's reasoning and was prompted only with the taxonomy definitions; however such a mitigation is not perfect and the full set of judge adjudications is provided in Supplementary Information, Section 4. The majority of flagged conflicts (81.2\%) involved classes with substantial chemical overlap - for instance, \emph{Amination of Heteroaryl Halides} and \emph{Nucleophilic Aromatic Substitution}. The remaining 102 of 600 flagged reactions (17.0\%) involved mechanistically distinct classes and represent genuine failures of the classifier. Adjusting for boundary ambiguity yields an estimated true mislabelling rate of approximately 0.33\% across the full test set, comparable to the 0.59\% obtained for NameRXN by the same method. Thus indicating that the fully automated pipeline may achieve label reliability on par with human expert curation.

\subsubsection{Cross-taxonomy agreement with NameRXN}

The cleanlab analysis above estimates internal label consistency
but does not address whether the LLM-derived labels cluster reactions in a way similar to that of an expert-designed taxonomy. We use NameRXN, an expert-curated classification, as our reference, and ask how closely the two taxonomies agree on the same reactions. As the label sets share no common vocabulary, a conventional confusion matrix is unavailable, so we measured agreement through two complementary procedures. Greedy modal mapping was performed in both directions, for each NameRXN class taking the modal LLM class among its reactions and vice versa, yielding two mapping functions whose accuracies lower-bound the recoverability of each taxonomy from the other. We also computed adjusted mutual information (AMI), which compares partitions of unequal granularity without a shared label space and corrects for chance agreement; the closely related V-measure is tabulated alongside.

We repeated this analysis at successive depths of the LLM hierarchy to test how cross-taxonomy agreement varies with granularity (Table~\ref{tab:cross_taxonomy}). The AMI remains essentially flat, moving from 0.837 at L3 to 0.849 at L4 and 0.842 at L5; the partitions stay mutually informative at every depth. This flatness is the signature expected when two taxonomies describe the same chemistry at different resolutions.

The directional taxonomy to taxonomy mapping accuracies show how that agreement is structured, and how the structure changes with depth. At L3, where the LLM taxonomy is coarser
than NameRXN, NRX$\rightarrow$LLM accuracy is high (85.97\%)
because multiple NameRXN classes collapse cleanly onto single LLM
parents; for example, NameRXN's separate \emph{bromo}, \emph{chloro},
and \emph{iodo N-alkylation} entries all map to a single LLM class
covering N-alkylation. The reverse direction,
LLM$\rightarrow$NRX, is correspondingly lower (70.23\%) because the
LLM class contains reactions that NameRXN partitions further.
Moving to L4 and L5, the asymmetry inverts: NRX$\rightarrow$LLM
falls to 56.27\% at L5 while LLM$\rightarrow$NRX rises to 91.56\%.
The deeper LLM classes are finer than NameRXN's, with each one contained almost entirely within a single NameRXN class (high LLM$\rightarrow$NRX), but each NameRXN class spreads across
several LLM children (low NRX$\rightarrow$LLM).

These results indicate that the LLM-derived hierarchy is
not in disagreement with NameRXN but is describing the same
chemistry at a finer resolution. Where
NameRXN distinguishes N-alkylation by leaving group at its
finest level, the LLM hierarchy reaches that distinction at L4
and continues to subdivide at L5 into reagent- and
condition-specific variants that NameRXN does not represent.

We then applied the same comparison to Rxn-INSIGHT~\citep{rxn-insight},
the leading open-source reaction classification tool, to test
whether the patterns observed against NameRXN generalise to
alternative reference taxonomies. The directional accuracies
follow the same inverting pattern as for NameRXN, with
LLM$\rightarrow$Rxn-INSIGHT rising from 71.6\% at L3 to 93.8\%
at L5 as the LLM hierarchy becomes successively finer than the
reference partition. However, AMI behaves qualitatively
differently: against NameRXN, AMI is essentially flat across
depths (0.837 to 0.849 to 0.842), whereas against Rxn-INSIGHT
it declines monotonically (0.724 to 0.720 to 0.691). A declining AMI indicates that the finer LLM partitions are no longer recoverable from Rxn-INSIGHT at all; since the same partitions remain recoverable from NameRXN, this reflects the limited resolution of Rxn-INSIGHT's smaller template set rather than spurious structure in the LLM hierarchy.

Together, these comparisons indicate that the LLM hierarchy agrees with expert-curated chemistry at matched resolution and resolves reactions more finely than either reference.

\begin{table}[ht]
\centering
\caption{Cross-taxonomy agreement between the LLM-derived hierarchy and two reference taxonomies (NameRXN, proprietary; Rxn-INSIGHT, open-source), at successive depths of the LLM hierarchy. Greedy accuracies measure recoverability in each direction; AMI and V-measure compare partitions without requiring a shared label space. Train fold (538k reactions); test-fold numbers in parentheses.}
\label{tab:cross_taxonomy}
\begin{tabular}{lcccccc}
\toprule
LLM depth & Ref$\rightarrow$LLM & LLM$\rightarrow$Ref & AMI & V \\
\midrule
\multicolumn{5}{l}{\emph{LLM vs NameRXN}} \\
L3 & 85.97\% (85.32) & 70.23\% (69.79) & 0.837 (0.810) & 0.842 (0.840) \\
L4 & 64.92\% (63.92) & 85.09\% (85.43) & 0.849 (0.810) & 0.858 (0.858) \\
L5 & 56.27\% (55.03) & 91.56\% (92.14) & 0.842 (0.793) & 0.854 (0.854) \\
\midrule
\multicolumn{5}{l}{\emph{LLM vs Rxn-INSIGHT}} \\
L3 & 67.34\% (65.40) & 71.56\% (71.88) & 0.724 (0.683) & 0.729 (0.713) \\
L4 & 52.94\% (50.85) & 90.86\% (91.28) & 0.720 (0.653) & 0.730 (0.712) \\
L5 & 45.67\% (43.75) & 93.81\% (94.32) & 0.691 (0.603) & 0.706 (0.688) \\
\bottomrule
\end{tabular}
\end{table}
\subsection{Autonomous Reaction Template Generation}

\begin{figure}[htb]
    \centering
    \includegraphics[width=0.95\textwidth]{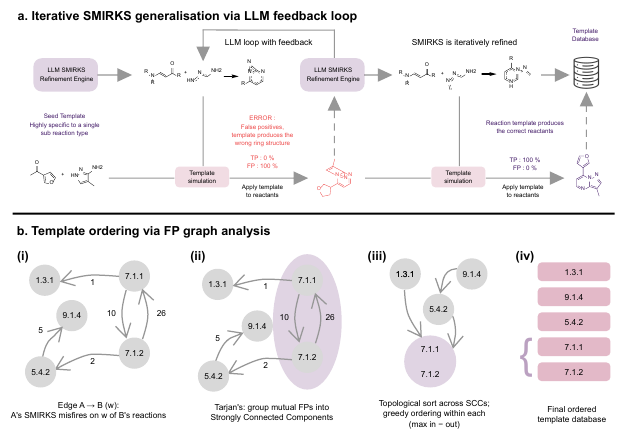}
    \caption{In \textbf{a.} we demonstrate a worked example of the template generalisation approach for pyrazolo[1,5-a]pyrimidine synthesis. The coloured molecules after the bottom arrows are the result of applying the generated template to the reactants. In \textbf{b.} we highlight a scheme demonstrating the template ordering. in (i) the value \textit{n} assigned to an edge from A to B indicates that \textit{n} templates of Class A produce a false positive on class B.}
    \label{fig:template_parat}
\end{figure}
Expert-curated synthesis and reaction classification tools  such as Synthia and Rxn-INSIGHT, respectively, rely on the manual authoring of reaction SMIRKS, a labor-intensive bottleneck. In raw data corpora, a single transformation category often manifests as dozens of overly specific, templates due to artefacts of the algorithmic rules used for extraction. Distilling these into a minimal set of generalised SMIRKS simplifies the computational representation and provides robust, broad coverage at intermediate taxonomic levels. An obvious choice for this task is am LLM, however while LLMs possess broad chemical knowledge, their ability to author syntactically valid and chemically precise SMIRKS remains unproven. Successfully automating this curation process would demonstrate that LLMs can reliably translate high-level chemical intent into rigid, deterministic rules, effectively automating a major cheminformatics workload.

To achieve this, we developed an iterative, LLM-driven generalization pipeline. To mitigate the over-generalization inherent in purely generative approaches, we implemented a five-stage validated learning loop. Each reaction class was partitioned into an 80/20 train/test split. The LLM generated candidate SMIRKS using the training set, which were subsequently screened against the broader corpus of different reaction classes to identify false positives. These false positives were fed back into an autonomous refinement phase, where the LLM iteratively modifies the SMIRKS patterns in response to qualitative accuracy and false positive metrics and qualitative examples of failure cases.

We validated this pipeline on a filtered subset of 665,901 reactions, requiring a minimum of five reactions per template. Performance on the held-out test set was evaluated at several different hierarchy levels.

\subsection{SMIRKS Database Ordering for First-Match Classification}

The generalised SMIRKS patterns enable fast, deterministic reaction classification via a first-match-wins strategy: given a query reaction, the database is traversed in order and the first SMIRKS that fires correctly determines the predicted class. However, when class A's SMIRKS also fires correctly on class B's reactions (a false positive), the ordering determines which class is assigned. To reduce false positives, we use an ordering algorithm detailed in section  \ref{sec:methods:ordering}. Testing 665,901 reactions against all 4,964 SMIRKS patterns revealed the complete false-positive graph; the ordering algorithm eliminated 95.8\% of false positives, with the remaining 4.2\% arising from unavoidable mutual relationships within 215 strongly connected components (SCCs). The end-to-end classification accuracy of the ordered database, alongside its hybrid MDP-gated counterpart, is evaluated against the held-out test fold and an out-of-distribution academic corpus in Section \ref{sec:detclassifier}.

\subsection{A deterministic reaction classifier}\label{sec:detclassifier}

\begin{table}[ht]
\centering
\caption{In-distribution accuracy by hierarchical tier on two held-out USPTO splits. Each cell reports \emph{overall} accuracy (correct over all test reactions) with \emph{covered} accuracy (correct over reactions on which a template fired) in parentheses. \emph{Templates} is the size of the template library each mode draws on. Split~A is frequency-weighted (per-template $80/20$); Split~B is template-balanced (one held-out reaction per template). Generalised SMIRKS is shown under the false-positive-minimising order and, for reference, under an arbitrary order of the identical $4{,}964$-pattern set (mean of three random shuffles), which isolates the effect of ordering. A correct prediction equals the ground-truth code truncated to the given tier.}
\label{tab:detclassifier_indist}
\begin{tabular}{lrcccc}
\toprule
Classifier & Templates & T2 & T3 & T4 & T5 \\
\midrule
\multicolumn{6}{l}{\emph{Split A --- frequency-weighted, $n=127{,}282$}} \\
Hybrid strict                & $44{,}449$ & 98.8 (99.6) & \textbf{97.9} (98.7) & 87.3 (88.5) & 83.4 (87.1) \\
Generalised SMIRKS, ordered  & $4{,}964$  & 92.1 (96.5) & 85.9 (90.1) & 58.7 (81.6) & 39.7 (78.3) \\
\quad arbitrary order        & $4{,}964$  & 82.9 (86.8) & 67.2 (70.5) & 40.7 (48.5) & 27.0 (41.1) \\
\addlinespace
\multicolumn{6}{l}{\emph{Split B --- template-balanced, $n=17{,}097$}} \\
Hybrid strict                & $44{,}449$ & 95.5 (99.1) & \textbf{94.1} (97.5) & 84.4 (88.0) & 76.9 (83.4) \\
Generalised SMIRKS, ordered  & $4{,}964$  & 82.9 (95.6) & 75.0 (86.8) & 55.4 (77.1) & 38.7 (68.5) \\
\quad arbitrary order        & $4{,}964$  & 78.7 (90.8) & 66.9 (77.5) & 46.8 (60.9) & 33.4 (54.3) \\
\bottomrule
\end{tabular}
\end{table}

The LLM-derived hierarchy, the generalised SMIRKS database, and the
false-positive-aware ordering algorithm developed in the preceding
subsections were assembled into a single classification system, ReactionClassifier. The system supports two inference
modes that lie at opposite ends of a granularity--robustness
trade-off. In the first, termed \emph{Hybrid strict}, a lightweight
multilayer perceptron trained on differential reaction fingerprints
(MDP) predicts a tier-3 class for the
query reaction. This prediction restricts the matching procedure to
the corresponding class-specific subset of the $44{,}449$ exact
reaction templates, and the predicted
label is accepted only when at least one template within the subset
fires on the input reactants and produces the correct ground truth product. The subsetting step removes many false positives associated with global first-match
strategies and preserves the deepest hierarchical detail represented
in the template library. In the second mode, \emph{Generalised
SMIRKS}, the MDP gate is omitted and the database of $4{,}964$
generalised SMIRKS is applied in the global
false-positive-minimising order of Section~\ref{sec:methods:ordering},
returning the first class whose SMIRKS matches the reaction.
Generalised SMIRKS does not reach the deepest
tiers but is less sensitive to gate error and to deviations from the
training distribution of the MLP. The MDP prediction is used only as
a gate; standalone MDP classification accuracies are not reported, as
the prediction is probabilistic and hence does not provide a strong guarantee of correctness

Classification accuracy depends on how the held-out set is constructed; rather than treating a single test distribution as definitive, we report two complementary splits of the labelled USPTO reactions ($628{,}870$ reactions). 
In the \emph{frequency-weighted} split (Split~A), each retrosynthetic template contributes $80\%$ of its reactions to training and $20\%$ to the test fold, so that common transformations dominate the test set in proportion to their natural abundance ($n=127{,}282$), establishing a baseline for everyday synthetic utility. 
In the \emph{template-balanced} split (Split~B), a single reaction is held out per template, weighting every template equally and over-representing the long tail of rare transformations ($n=17{,}097$) to stress-test the classifier's absolute breadth. 
Both splits retain the remaining reactions of each template in the training fold, and a separate MDP gate was trained on each split's training fold; the two splits therefore measure accuracy on novel substrates of templates seen during training rather than extrapolation to unseen chemistry, the latter being assessed on the out-of-distribution corpora below. 

We report two metrics (Table~\ref{tab:detclassifier_indist}).
\emph{Overall accuracy} is the fraction of all test reactions assigned
the correct code at a given tier, while \emph{covered accuracy}
restricts this fraction to reactions on which at least one template
fired and therefore measures reliability once the classifier commits to
a label. Hybrid strict reaches $97.9\%$ overall at tier~3 on the
frequency-weighted split and $94.1\%$ on the template-balanced split, at
a median of $6$--$7$~ms per reaction; its covered accuracy is higher
still ($98.7\%$ and $97.5\%$), so that a committed classification is
rarely wrong and the residual error is dominated by abstention rather
than misclassification. Generalised SMIRKS is lower overall ($85.9\%$
and $75.0\%$ at tier~3) because, without the neural gate, fewer
reactions are covered and a pattern ordered ahead of the correct class
can fire on reactions the gate would otherwise have routed elsewhere;
its covered accuracy at tier~3 ($90.1\%$ and $86.8\%$) is correspondingly
closer to that of Hybrid strict. Accuracy declines at tiers~4 and~5 for
both modes, where classes are distinguished by reagent and condition
information that substrate structure alone does not determine.

The ordering of the generalised SMIRKS database is itself a substantial
contributor to classifier performance. Applying the identical $4{,}964$-pattern set in
an arbitrary order lowers tier-3 covered accuracy from $90.1\%$ to
$70.5\%$ on Split~A and from $86.8\%$ to $77.5\%$ on Split~B, with the
gap widening at deeper tiers. Because coverage is unchanged by
reordering, this difference reflects only which class is returned when
several patterns match a reaction: the false-positive-minimising order
recovers roughly $10$--$20$ percentage points of coverage-conditional
accuracy compared to arbitrary ordering.

\subsection{Out-of-distribution coverage}\label{sec:extensibility}

The held-out evaluation above is an in-distribution test: every
reaction belongs to a template that the LLM pipeline labelled, so the
two splits measure internal consistency on novel substrates of known
templates rather than true generalisation. We therefore evaluate on two out-of-distribution (OOD) test sets, benchmarking against the established tools NameRXN
(proprietary) and Rxn-INSIGHT~\citep{rxn-insight} (open-source)
(Table~\ref{tab:detclassifier_latency}).

CRD-2025 is a single-reaction-centre subset of reactions reported in
academic publications dated 2025 and later ($n=9{,}296$) \cite{chemicalreactiondatabase}. It is
out-of-distribution along two axes: temporally, it lies entirely beyond
the 1976--2016 USPTO window, and chemically, it reflects the priorities
of academic methodology development rather than the process- and
scale-oriented chemistry that dominates the patent record. As no ground-truth labels are available for these reactions, we report only coverage (the
fraction assigned a non-null label) and latency. For the second test set, we choose ring forming reactions (RingBreaker) derived from the AiZynthTrain retrosynthesis pipeline~\citep{3aizynthtrain}. Ring forming reactions offer an interesting area for our approaches - due to complex stereo and regio chemical constraints, retrosynthesis tools can struggle to apply known ring forming reactions to novel chemical contexts, and hence the ability of a SMIRKS set to model such chemistry indicates how well they have been generalised. We construct a set comprising (p$n=57{,}234$)  ring-forming reactions from the USPTO corpus for this test.

\begin{table}[ht]
\centering
\caption{Coverage and median per-reaction inference latency on two out-of-distribution corpora. CRD-2025: single-reaction-centre academic reactions published in 2025 and later ($n=9{,}296$). RingBreaker: ring-forming USPTO reactions that existing retrosynthesis pipelines were unable to classify ($n=57{,}234$). Latency measured under 16-core multiprocessed inference on a workstation CPU.}
\label{tab:detclassifier_latency}
\begin{tabular}{lcccc}
\toprule
 & \multicolumn{2}{c}{CRD-2025 ($n=9{,}296$)} & \multicolumn{2}{c}{RingBreaker ($n=57{,}234$)} \\
\cmidrule(lr){2-3}\cmidrule(lr){4-5}
Method & Coverage (\%) & Median (ms) & Coverage (\%) & Median (ms) \\
\midrule
Hybrid strict   & 64.0 &  5.88 & \textbf{80.6} & 17.81 \\
Generalised SMIRKS  & 68.3 &  9.28 & 77.3          & 34.28 \\
NameRXN         & \textbf{89.3} &  3.09 & 73.7  &  4.64 \\
Rxn-INSIGHT     & 64.8 & 12.25 &  6.9          & 16.64 \\
\bottomrule
\end{tabular}
\end{table}

On the academic corpus, NameRXN attains the highest coverage
($89.3\%$), reflecting the long-tail entries accumulated in its rule
library over more than a decade of manual
curation~\citep{schneider2016big} (Supplementary
Figure ?? reports the per-template coverage
distribution). Among the data-derived methods, however, Generalised
SMIRKS attains the highest coverage ($68.3\%$), above both Hybrid
strict ($64.0\%$) and Rxn-INSIGHT ($64.8\%$). This is direct evidence
of generalisation: Generalised SMIRKS draws on only $4{,}964$ patterns, roughly a tenth of the $44{,}449$ exact templates available to
Hybrid strict, yet reaches more of the novel academic chemistry,
because abstracting away substrate-specific detail lets a compact rule
set match transformations generalise well to a greater diversity of substrates. The effect is most pronounced on the ring-forming
corpus, where our modes ($80.6\%$ and $77.3\%$) exceed
NameRXN ($73.7\%$) and far exceed Rxn-INSIGHT ($6.9\%$, whose rule set
is dominated by acyclic functional-group transformations). As ring
formation can be the limiting step in retrosynthetic planning, an
$80\%$ classifiable fraction approaches complete coverage of the
template space encountered in single-step retrosynthesis. Both modes
operate at a median of $6$--$9$~ms per reaction, roughly twofold faster
than Rxn-INSIGHT, while assigning labels at three additional taxonomic
tiers (Table~\ref{tab:cross_taxonomy}).

For both deterministic modes the primary failure mode on out-of-distribution
input is abstention: a reaction for which
no template fires is returned unclassified, so the determinism guarantee of the symbolic layer
is preserved. The roughly one-third of CRD-2025 reactions on which the
deterministic layer abstains corresponds to transformations that fall
outside the existing hierarchy, for which no generalised or specific SMIRKS matches
the recorded product. The following subsection applies the multi-agent
pipeline of Section~\ref{sec:methods:pipeline} to the remaining data.

\subsection{LLM fallback on unclassified academic chemistry}\label{sec:fallback}

The reactions on which the deterministic layer abstains are not a random
sample of chemical space. We hypothesise that such failures likely fall into categories of chemistry that are either novel or poorly represented in industrial process chemistry. Adapting a taxonomy to such reactions has conventionally
required extensive manual curation, as in fixed tools such as NameRXN and
Rxn-INSIGHT. The two-layer architecture automates this step: reactions that the
symbolic layer cannot place are routed to the LLM pipeline, which classifies
them and extends the taxonomy on demand.

We then quantified how much of this coverage reflected assignment to
existing classes rather than expansion of the taxonomy. We note the template database does not necessarily cover all reactions that could be placed into the existing taxonomy; coverage records
only whether a reaction received a label, not whether that label already
existed in the hierarchy. As the pipeline can create new classes as
well as assign existing ones, the fraction of cohorts routed to the
generator agent measures how far the academic corpus extends beyond the
patent-derived hierarchy.

\paragraph{Classification outcome.}
  A final classification was assigned to $2{,}911$ of the $2{,}990$
  processed single-center reactions ($97.4\%$); the remaining $79$
  ($2.6\%$) did not receive a final label and are candidates for a
  targeted rerun. Combined with the $64.0\%$ coverage of Hybrid strict on
  the parent CRD-2025 corpus, the two-layer architecture classifies
  $8{,}861$ of the $9{,}296$ reactions ($95.3\%$), exceeding the $89.3\%$
  coverage of NameRXN on the same corpus without recourse to manually
  authored rules for post-2016 academic chemistry.

  \paragraph{Taxonomy growth.}
  The fallback was applied both to the single-center abstention set and to
  a further $960$ multicenter reactions that are excluded from the symbolic
  classifier's evaluation. Across both sets, a total of $1{,}942$ entries
  were appended to the hierarchy (Table~\ref{tab:taxonomy_growth}). Of
  these, $135$ were located at the third hierarchical tier, the level of
  named reaction types such as \emph{N-arylation with aryl halides} or
  \emph{Suzuki--Miyaura coupling}; the remaining additions were
  distributed across deeper tiers, with $687$ at L4, $836$ at L5, and
  $284$ at the condition-level (L6/L7) tiers. Of the $3{,}827$ cohorts
  processed, $37.0\%$ were routed to the generator agent and therefore
  required the creation of a new taxonomy entry rather than assignment to
  an existing class. The magnitude of this fraction is consistent with the
  interpretation that the academic literature sampled in CRD-2025 occupies
  a region of chemical space distinct from that represented in the patent
  corpus on which the hierarchy was originally generalised.

  \begin{table}[htbp]
  \centering
  \caption{Hierarchy entries added by the LLM fallback on the CRD-2025
    single-center and multicenter abstention sets.}
  \label{tab:taxonomy_growth}
  \small
  \begin{tabular}{@{}lrl@{}}
  \toprule
  Hierarchy level & New entries & Description \\
  \midrule
  L3 (reaction type)       & 135  & Named reaction classes \\
  L4 (reaction subtype)    & 687  & Substrate-specific variants \\
  L5 (sub-subtype)         & 836  & Mechanistic distinctions \\
  L6/L7 & 284  & Reagent/condition variants \\
  \midrule
  Total                    & $1{,}942$ & \\
  \bottomrule
  \end{tabular}
  \end{table}

\paragraph{Origin of the new taxonomy entries.}
The $135$ L3 entries introduced during the fallback correspond to
established synthetic methodologies that are underrepresented in the
USPTO corpus, rather than to transformations first reported in 2025.
The $\alpha$-selenenylation of carbonyl compounds is one such case:
introduced in the 1970s and routinely employed in academic total
synthesis as the entry step into the selenoxide elimination sequence to
$\alpha,\beta$-unsaturated carbonyls, the reaction is largely absent
from pharmaceutical patent corpora, a pattern consistent with the
regulated status of selenium as an elemental impurity in drug substances
 \citep{ich2022q3dr2,santoro2014green}. The pre-fallback taxonomy contained selenium
chemistry only at the SeO\textsubscript{2}/Riley oxidation and
selenoxide elimination stages; the upstream C-Se bond-forming step was
added during the present run as a new L3 class with three L4 substrate
variants. The amination of aryl nonaflates by secondary aliphatic
amines provides a second case. Nonaflate substrates were already
represented in the taxonomy in the context of Suzuki-Miyaura coupling
and the amination of heteroaryl nonaflates; the corresponding
aryl-amination class, which appears more frequently in academic
methodology work than in pharmaceutical process chemistry, was not, and
was added on demand. In both cases the taxonomy expansion reflects a
distributional difference between the patent corpus on which the
hierarchy was generalised and the academic corpus on which it was
subsequently applied, rather than the emergence of new chemistry. Full
hierarchical pathways for both examples are provided in Supplementary Information

\paragraph{C--H functionalisation and reductive cross-coupling.}

  Beyond isolated substrate classes, the entries introduced
  during the fallback include a coherent set of recognisable, well-established
  synthetic methodologies. The two largest families are direct
  C--H functionalisation, which appears in $23$ of $298$ source DOI's ($7.7\%$), and reductive or radical cross-coupling, which
  appears in $21$ ($7.0\%$). For C--H functionalisation, the emergent classes include the direct, halide-free C--H borylation of (hetero)arenes (the Ishiyama--Miyaura--Hartwig reaction), directed C(sp\textsuperscript{3})--H arylation of amides , and intramolecular carbene C--H insertion of diazo compounds . For reductive cross-coupling, they include the decarboxylative coupling of $N$-hydroxyphthalimide (redox-active) esters including the construction of bicyclo[1.1.1]pentane bioisosteres  and the nickel-catalysed reductive cross-electrophile coupling of alkenyl and alkyl halides. The classes generated are identifiable, named methodologies rather than ad hoc groupings, an indication that the fallback recovers meaningful synthetic chemistry absent in the USPTO corpus.

  \subsection{Expert-chemist validation of classification accuracy}\label{sec:expert_validation}
  \begin{table}[htbp]
  \centering
  \caption{Expert-chemist adjudication of reaction classifications, comparing the
  LLM-derived taxonomy against NameRXN across two evaluation pools. Equal-weighted
  across three chemists: each of the 69 reactions was graded by all three raters
  (207 gradings); percentages are the mean across raters. Inter-rater agreement
  was moderate (Fleiss $\kappa \approx 0.40$ for the LLM labels, $0.45$ for NameRXN).}
  \label{tab:human_eval}
  \begin{tabular}{lccccc}
  \toprule
  System \& Pool & $n$ & Correct & Acceptable / Imprecise & Wrong & N/A \\
  \midrule
  \textbf{LLM Taxonomy (All)} & \textbf{69} & \textbf{82.6\%} & \textbf{6.3\%} & \textbf{8.2\%} & \textbf{2.9\%} \\
  $\hookrightarrow$ Original pool & 47 & 90.1\% & 5.7\% & 4.3\% & 0.0\% \\
  $\hookrightarrow$ New pool & 22 & 66.7\% & 7.6\% & 16.7\% & 9.1\% \\
  \midrule
  \textbf{NameRXN (All)} & \textbf{69} & \textbf{61.8\%} & \textbf{17.9\%} & \textbf{7.2\%} & \textbf{13.0\%} \\
  \bottomrule
  \end{tabular}
  \end{table}
  To benchmark the taxonomy against the \textit{de facto} standard, NameRXN, we
  conducted a blind, dual-label expert evaluation by three chemists on a common
  stratified sample of the out-of-distribution CRD-2025 corpus ($n=69$ reactions,
  each graded by all three raters). Raters graded blinded, randomised label pairs
  as \textit{Correct}, \textit{Acceptable but imprecise}, or \textit{Wrong}, and
  all statistics are weighted equally across raters. The results in
  Table~\ref{tab:human_eval} indicate that the LLM-derived hierarchy achieves
  absolute correctness comparable to NameRXN, with a statistically insignificant
  difference in outright error rates ($8.2\%$ vs.\ $7.2\%$, respectively). Neither
  system is meaningfully more prone to absolute misclassification.

  Turning to label precision, the LLM labels were graded fully \textit{Correct} in
  $82.6\%$ of cases, whereas NameRXN achieved a \textit{Correct} rating in only
  $61.8\%$ of cases, with $17.9\%$ of its labels penalised as \textit{Acceptable
  but imprecise}. In paired head-to-head assessments, the LLM assignment was
  strictly preferred over NameRXN in $47$ out of $61$ discordant reactions. This
  demonstrates that the primary failure mode of fixed taxonomies relative to human
  preference is under-specification rather than error.

  We also observed a notable performance variance between the two sampling pools.
  LLM assignments for reactions covered by rules extracted from the USPTO corpus
  were graded \textit{Correct} in $90.1\%$ of cases, compared to $66.7\%$ for
  previously unclassified reactions routed to the LLM generator. We propose several
  hypotheses for this gap: original-pool labels used cohort-based regularisation in
  the naming process (up to five reactions per template), which stabilises the
  LLM's output, whereas fallback labels were assigned from a single reaction; the
  academic methodology literature over-represented in the ``new'' pool features
  highly specific or unusual transformations that are inherently more ambiguous to
  classify; and the common USPTO reactions comprising the ``original'' pool are
  likely better represented in the LLM's pre-training corpus than the novel
  academic reactions in the fallback pool.

\section{Conclusion}
Recent work has established that LLMs display a strong understanding of the language of chemistry; here we demonstrate that they can also write its grammar, performing at scale a detail-orientated and labour intensive task previously the preserve of cheminformaticians. What previously required years of expert curation can now be performed on demand at costs affordable to academic research laboratories.

Reaction classification poses a dual challenge; assignments must be deterministic and reliable, while the underlying taxonomy should be dynamically adaptable to the creative research which defines the frontier of organic chemistry. The architecture presented here resolves this tension by separating the two tasks; a deterministic symbolic layer, realised as a database of named reaction SMIRKS, classifies known chemistry quickly, while a generative language-model layer is invoked on demand to categorise novel transformations and propose new taxonomic entries where required. This pattern need not be specific to chemistry; any domain in which a long-tailed, evolving body of knowledge must be both deterministically applied and continuously expanded could benefit from the same abstract architecture.

The taxonomy resolves reactions more finely than existing classifiers
(Table~\ref{tab:cross_taxonomy}), which is useful both to chemists and to
downstream methods. For the former, reaction collections can be searched and
filtered at the level of specific named transformations rather than broad
categories. For the latter, the finer resolution gives class-conditioned generative methods more specific control. Synthesisability-constrained generation can be steered only as precisely as its reaction labels allow. Rather than being restricted to existing patent data, the ability to generate rules on demand allows generative models to be conditioned directly on novel chemistry, enabling the exploration and mapping of the chemical space accessible via newly discovered transformations. Finally, the taxonomy is not a fixed artefact: local aliases, proprietary chemistry, and institution-specific nomenclature can be added, enabling cheap customisation to proprietary preferences.

The pipeline presented here is tuned to the USPTO corpus, which over-represents pharmaceutical chemistry and under-represents organometallic, materials, and process-scale transformations; extending it to specialised reaction corpora, such as those covering electrochemistry, flow chemistry, or biocatalysis, will require additional curation effort, although the underlying methodology transfers directly. The pipeline depends on a reasonable seed hierarchy to bootstrap from, and while it grows that hierarchy organically, it cannot reliably construct one from nothing. Looking further, the present work is fundamentally retrospective; it organises and generalises chemistry that has already been recorded. Preliminary results suggest that language models can also move from curating known reactions to proposing transformations absent from any existing corpus, and we are pursuing this in ongoing work. Together, curation and generation suggest a route toward synthesis planners that are no longer bounded by the coverage of their training data and would enable CASP tools to begin approaching the synthesis of complex molecules whose synthesis currently requires bespoke methodology development.

\section{Methods}\label{sec:methods}

\subsection{Dataset and preprocessing}\label{sec:methods:data}

\subsubsection{Working dataset}
Our starting point was the USPTO reaction corpus of \citep{Lowe2012, Lowe2017}, comprising approximately
1.8~million chemical reactions extracted from US patent filings between
1976 and 2016, encoded as SMILES strings of the form
\texttt{reactants>reagents>products}. The corpus was processed with the AiZynthTrain pipeline \citep{3aizynthtrain}.

We further restricted the dataset to reactions whose templates occur at least
three times in the filtered corpus, providing the minimal redundancy
required for cohort-level analysis and downstream template
generalisation. After these filters, 860{,}675 reactions covering
42{,}125 distinct templates remained and constitute the working
dataset throughout this study.

\subsubsection{Cohort grouping by template hash}
Reactions sharing an identical template hash undergo the same core
structural transformation by construction. We exploited this
redundancy by grouping such reactions into \emph{cohorts} that received
a single, jointly-determined classification label. From each cohort we
drew up to five reactions (or all reactions, when fewer than five were
available) for inclusion in the LLM prompts; this provides the agents
with substrate diversity and functions as a form of regularisation to noise present in individual reaction samples. In aggregate, 179{,}495 reactions across the 42{,}125 cohorts were sampled and submitted to the language-model pipeline, with each cohort's assigned
label subsequently broadcast to every reaction sharing its template
hash.

\subsubsection{Stratified train/test split}
For SMIRKS template generation and held-out evaluation we performed an
80/20 stratified split at the finest available tier of the LLM-derived
hierarchy, using a fixed random seed of $210{,}995$ for
reproducibility. Class-level filtering criteria specific to template
generalisation, including the minimum-class-size threshold, are
described together with the SMIRKS pipeline in
Section~\ref{sec:methods:smirks}.
\subsubsection{Out-of-distribution evaluation corpora}\label{sec:methods:ood}

To probe generalisation beyond the USPTO timeframe and beyond
acyclic chemistry, we assembled two evaluation corpora that are
disjoint from the training data.

\paragraph{Chemical Reaction Database (CRD-2025).}
Reactions extracted from chemistry publications dated 2025 and
later were scraped from the publicly maintained Chemical Reaction Database (CRD)
(\url{https://kmt.vander-lingen.nl}), yielding an initial raw scrape size of 11,088 reactions.
Records were canonicalised using RDKit~\citep{rdkit} and atom-mapped with
RXNMapper~\citep{schwaller2021extraction}. We subsequently applied a
\emph{single-reaction-centre} filter, retaining only reactions for which the mapper
assigned a single connected component of changed atoms and for which every
reactant-side molecule contributed at least one heavy atom to the product. This pipeline
removes multi-stage transcriptions, charge-transfer artefacts, and misplaced reagents,
leaving a final evaluation set of 9,296 reactions.

\paragraph{RingBreaker template library.}
The second OOD corpus was the USPTO RingBreaker template library
distributed with AiZynthTrain \citep{3aizynthtrain,thakkar2020ring}. This set
comprises 57,234 ring-forming reactions retained on the criterion that the
underlying retrosynthetic template occurs at least three times across the full
USPTO corpus. This dataset isolates ring-forming chemistry that
demands complex stereochemical and regiochemical constraints.

\subsection{Multi-agent LLM classification pipeline}\label{sec:methods:pipeline}

\subsubsection{Architecture overview}
We decomposed reaction classification into a sequence of five
specialised language-model agents (Figure~\ref{fig:classification_main}).
Each cohort, defined in Section~\ref{sec:methods:data}, is passed
through the pipeline in turn: a hierarchy agent assigns a coarse
two-level code, a detailed agent refines it to the deepest applicable
tier, a verifier agent audits the assignment, a generator agent
proposes a new taxonomy entry whenever verification fails, and an
aggregator agent reconciles the new proposals against the existing
hierarchy. This structure follows the Chain-of-Verification
paradigm of Dhuliawala et al.~\citep{dhuliawala2024chain}, in which
separating proposal from verification reduces the propagation of
self-consistent but factually incorrect claims. Throughout the run,
the taxonomy itself is treated as an expandable object: the aggregator
appends new entries to a dynamic mapping that all subsequent agent
calls consult, so the hierarchy grows organically as the pipeline
encounters chemistry that is not represented in the seed taxonomy.

\subsubsection{Hierarchy agent}
The hierarchy agent assigns each cohort to one of 68 fixed
sub-classes drawn from the RXNO ontology~\citep{rxno} and the
analyses of Carey et al.~\citep{carey2006analysis} and Roughley et
al.~\citep{roughley2011medicinal}. The complete two-level seed
taxonomy is embedded in the system prompt as nine super-classes
spanning heteroatom alkylation, acylation, C–C bond formation,
heterocycle formation, protections, reductions, oxidations,
functional-group interconversion, and functional-group addition. The
agent receives the cohort's reaction SMILES and the shared
retrosynthetic template, and returns a single
\verb|<reaction_class>X.Y</reaction_class>| tag. The level-1 and
level-2 codes are immutable across the run; later agents are not
permitted to alter them.

\subsubsection{Detailed agent}
The detailed agent refines the assignment to the deepest applicable
tier of a seven-level taxonomy. Given the L2 code from the previous
stage, the agent receives the corresponding subtree of the dynamic
mapping as context, together with the cohort SMILES and the shared
retrosynthetic template. Classification follows a strict two-path
algorithm. In Path~A, the agent first searches the hierarchy for a
class whose reactant functional groups exactly match those present
in the cohort, then optionally descends to a deeper reagent-specific
variant. In Path~B, applicable when no precise functional-group
match can be found, classification terminates immediately at the
nearest valid parent class with the suffix \texttt{.Other} appended.
The agent emits five XML tags encoding the level-1 through level-4+
labels and the complete hierarchical code. Levels 3 and 4 are
required to be derivable from reactants and products alone; reagent
identity is admitted as a discriminator only at level 5 and beyond.

\subsubsection{Verifier agent}
The verifier audits the proposed classification without access to
the detailed agent's reasoning, which removes the bias toward
reproducing the original assignment. The agent returns one of three
outcomes. A valid classification is reported as
\verb|<match>true</match>| with action \texttt{continue}. A standard
mismatch, in which the high-level category is correct but the
specific reactant or reagent identity is wrong, is reported with
\verb|<match>false</match>| and action \texttt{continue}, routing
the cohort to the generator. A fatal hierarchy error, in which the
proposed L1/L2 category is fundamentally incompatible with the
reaction (for example, an alkene reduction misplaced under
N-acylation), is reported with action
\texttt{incorrect\_hierarchy}. Cohorts flagged in this way are
excluded from further processing and labelled
\texttt{HIERARCHY\_MISMATCH} in the output, preventing downstream
code conflicts that would otherwise arise when chemically unrelated
reactions are forced into a shared parent class.

\subsubsection{Generator agent}
The generator proposes a new taxonomy entry for any cohort that
exits the verifier with action \texttt{continue} and
\verb|<match>false</match>|. The agent is instructed to follow a
``what versus how'' naming principle: levels 3 and 4 must describe
the core transformation in reagent-agnostic terms (the reactant
functional groups and the product functional groups), while levels
5 and deeper encode named reactions or generalisable condition sets
and are prefixed with \texttt{cond:} for downstream parsing.
Proposals are emitted as a single hierarchy line of the form
\verb|code label| together with any required new parent levels, a
suggested parent code, and a free-text rationale. The agent is
forbidden from proposing modifications to level-1 or level-2 codes.

\subsubsection{Aggregator agent}
Generator proposals are noisy at the cohort level: distinct cohorts
that describe the same novel transformation will independently
propose nearly identical entries with arbitrary placeholder codes.
The aggregator consolidates a batch of proposals, grouped by their
shared L2 parent, into a coherent set of additions. It deduplicates
chemically overlapping proposals, assigns sequential codes that fit the
existing hierarchy, and emits both a list of new hierarchy lines
and a JSON map from in-batch proposal indices to the final assigned
codes. The dynamic mapping is updated atomically after each
aggregator call, so cohorts processed later in the same run see and
reuse the new entries rather than re-proposing them. A code
conflict, defined as the same code already present in the mapping
under a different label, is recorded and the conflicting line is
skipped, with the affected cohort marked \texttt{CONFLICT:} for
post-hoc resolution.

\subsubsection{Models and inference}
Both Gemini~3 Flash and Gemini~3 Pro were used for inference,
accessed through the Google GenAI SDK. Lower-cost classification
stages were assigned to Gemini~3 Flash and the verification,
generation, and aggregation stages to Gemini~3 Pro, reflecting the
larger reasoning budget required to audit and extend the taxonomy.
All agents were run with a sampling temperature of $0.1$. Cohorts
were processed in batches with inter-cohort parallelism implemented
via a thread pool of up to eight workers. The pipeline is fully
resumable: intermediate stage artefacts are checkpointed after each
batch, allowing a run to be restarted from any stage without
re-issuing already-completed model calls.

\subsection{Label-noise estimation}\label{sec:methods:noise}

\subsubsection{Confident learning with out-of-fold predictions}
We estimated the residual annotation error in both the LLM and
NameRXN (version 3.7.0) label sets using the confident learning framework of
Northcutt et al.~\citep{northcutt2021confident}, implemented in the
\texttt{cleanlab} package. For each reaction we computed a
Morgan Difference Fingerprint
(MDP)  and trained a multilayer
perceptron classifier on the resulting fingerprint vectors. To
prevent the classifier from memorising potentially noisy labels,
predicted class probabilities were obtained by 5-fold stratified
cross-validation: each reaction was scored by a model instance
trained on a fold that excluded its own label. The cleanlab
self-confidence procedure was then applied to these
out-of-fold probabilities to flag examples whose original label
disagrees with the model's confident prediction.

\subsubsection{Hierarchy depth alignment}
\label{methods:hierarchy}
The LLM-derived hierarchy is finer than the NameRXN taxonomy by construction. Because finer classifications naturally incur higher baseline label noise, comparing the two raw outputs directly would conflate taxonomy resolution with classification accuracy. To enable a like-for-like comparison, we truncated the LLM labels to a depth yielding a class count comparable to the NameRXN classification over the identical reaction set (level 3 of the LLM hierarchy). This allowed for a direct statistical comparison of label noise without requiring explicit mapping between the two ontologies. All baseline labels were generated using NameRXN version 3.7.0; we note that newer versions of NameRXN contain a higher number of reaction classes, which would require truncation to a correspondingly deeper level.

\subsubsection{LLM-as-judge adjudication}
\label{methods:LLMasjudge}
Confident-learning noise estimates conflate genuine mislabelling
with ambiguity at taxonomy boundaries, so we further adjudicated
the flagged reactions in the LLM-labelled test set
($n = 30{,}802$). Of the 600 reactions flagged by cleanlab
($1.95\%$ of the test set), 321 exhibited a direct conflict
between the MLP's predicted class and the original LLM label and
were submitted to a separate language-model judge. The judge was
provided with the two class definitions only and was asked to
determine whether the classes share substantive mechanistic or
substrate overlap; it had no access to the original classifier's
reasoning. Both systems share broadly similar pre-training
corpora, so this adjudication is not strictly independent, but the
absence of the original chain-of-thought substantially reduces
direct contamination. The judge classified $81.2\%$ of the
adjudicated conflicts (261 of 321) as taxonomy-boundary
ambiguities. Across the full set of 600 cleanlab-flagged
reactions, $17.0\%$ (102 reactions) were judged mechanistically
distinct misclassifications, giving an estimated true mislabel
rate of approximately $0.33\%$ relative to the full test set.

\subsection{Autonomous SMIRKS generalisation}\label{sec:methods:smirks}

\subsubsection{Class filtering and template screening}
SMIRKS template generation requires several reactions per class to
support a reliable generalisation, so we restricted template
training to classes containing at least five reactions in the
training fold. This filter dropped 943 of the rarest classes,
accounting for approximately $15.1\%$ of reactions. For each
retained class, the most frequent
\texttt{TEMPLATE\_rr0rp1\_ring0} templates were selected by a
top-$N$ procedure that begins with $N=10$ and doubles until the
cumulative template coverage reaches 90\% or $N$ exceeds 50. The
shortest training-fold reaction matching each selected template
was retained as a worked example accompanying it in the prompt.

\subsubsection{Few-shot generalisation prompts}
The selected templates and worked examples for each class were
submitted to the language model together with a system prompt
that combines a SMARTS notation reference, six rules for
generalising over templates (atom-map handling, context-atom
abstraction, leaving-group merging, reaction-centre-only versus
contextual SMIRKS, splitting into multiple SMIRKS when the
reaction-centre transformation itself differs, and charge
handling), and a six-step reasoning procedure (identify the core
transformation, find reaction-centre atoms, check for distinct
patterns, identify variable positions, decide context level,
write and verify the SMIRKS). The prompt also includes three
worked examples covering the principal cases: an RC-only minimal
generalisation, halide-list merging with context generalisation,
and an instance requiring multiple SMIRKS for a single class.
Outputs were constrained to a Pydantic schema with two fields, a
free-text reasoning trace and a list of SMIRKS strings.

\subsubsection{Validation against worked examples}
Each returned SMIRKS was applied to the corresponding worked
examples using RDKit's \texttt{RunReactants}. To handle templates
with fewer reactant components than the recorded reaction, all
subsets and permutations of the reactant molecules were tried. A
SMIRKS was treated as correct on a given example if at least one
permutation produced a product that, after non-isomeric
canonicalisation, matched the recorded product; non-isomeric
comparison was used to avoid spurious failures arising from
differences in $E$/$Z$ encoding. The combined coverage of all
returned SMIRKS for a class was required to exceed $50\%$; if it
did not, the failing reaction together with the diagnostic
returned by RDKit was appended to the prompt and the model was
queried again, for up to three retry rounds.

\subsubsection{Cross-class false-positive testing}
After training, each SMIRKS was tested against the training-fold
reactions of every other class. A false positive was recorded
when a SMIRKS from class~A produced the correct product for a
reaction from class~B. Tests against reactions belonging to the
SMIRKS' own class, or to ancestor or descendant classes within
the hierarchy, were skipped, as were reactions with fewer
reactant components than the SMIRKS requires. Testing was
parallelised across eight worker processes via \texttt{joblib}
with reactions chunked at 200 per worker; each worker compiled
its own RDKit reaction objects locally because compiled reactions
cannot be transferred between processes. Each false-positive
record was annotated with the deepest shared ancestor of the
SMIRKS class and the true class, which allows false positives to
be categorised as same-tier-2 (closely related), same-tier-1
(within the same superclass), or cross-tier-1 (between unrelated
super-classes).

\subsubsection{Iterative refinement with rollback}
Classes with at least one false positive were re-prompted with a
refinement-specific system prompt that instructs the model to add
context-atom constraints, tighten hydrogen and degree descriptors,
restrict ring membership, or split overly broad patterns into
narrower ones. The user prompt included the current SMIRKS, up to
ten training examples that the refined SMIRKS must continue to
cover, and up to twenty false-positive examples grouped by true
class. After each round the refined SMIRKS was re-validated on
the training examples; if training-fold coverage dropped below
$80\%$ of the original recall, the refinement was rolled back and
the original SMIRKS retained. Otherwise the false-positive
examples were re-tested against the refined SMIRKS and any
eliminated false positives were removed from the working list.
Refinement proceeded for up to three rounds per class.

\subsubsection{Per-class tier-depth selection}
Different reaction classes are best described at different
hierarchical depths: well-defined transformations such as Suzuki
coupling admit a single broad SMIRKS at level~3, whereas finer
distinctions, for example condition-dependent variants of an
amidation, require level~4 or level~5 patterns. We therefore
pre-computed a recommended tier depth per class based on the
cumulative coverage of the top-$N$ templates at each level,
retaining the shallowest level at which $90\%$ template coverage
was attainable. The resulting mixed-depth SMIRKS database
contains 896, 1{,}498, and 1{,}360 classes generalised at
tier~3, tier~4, and tier~5 respectively.

\subsubsection{Held-out evaluation}
Final SMIRKS, refined where applicable and otherwise as
originally generated, were evaluated against the $20\%$ held-out
test fold. For each class, recall is the fraction of own-class
test reactions matched by at least one of its SMIRKS. Cross-class
false-positive testing was repeated on the test fold to obtain
per-class precision, specificity, $F_1$ score, and balanced
accuracy.

\subsection{SMIRKS database ordering}\label{sec:methods:ordering}

\subsubsection{First-match-wins inference}
Given the validated SMIRKS database, classification of an unseen
reaction proceeds by iterating the database in order and
returning the class of the first SMIRKS that fires correctly on
the reaction. This first-match-wins strategy is fully
deterministic but its accuracy depends on the database ordering:
when class~A's SMIRKS also fires correctly on reactions belonging
to class~B, the ordering of A and B determines whether the
reaction is correctly attributed to B or incorrectly attributed
to A.

\subsubsection{False-positive graph}
We modelled these inter-class dependencies as a weighted
directed graph in which nodes are reaction classes and an edge
$A \to B$ with weight $w$ records that class~A's SMIRKS fire
correctly on $w$ reactions whose true label is class~B. The
graph was built by testing every reaction in the database
against every SMIRKS, amounting to approximately
$3.3 \times 10^{9}$ individual SMIRKS applications across
665{,}675 reactions and 4{,}964 SMIRKS patterns. The procedure
mirrors the cross-class false-positive test used during
refinement (Section~\ref{sec:methods:smirks}) but is applied
without skipping ancestor or descendant relationships, so that
the resulting graph captures all SMIRKS-level interactions
present in the corpus.

\subsubsection{Ordering algorithm}
Finding a permutation that minimises the total weight of
remaining false positives is a weighted minimum-feedback-arc-set
problem and is NP-hard in general. We approached it in three
stages. First, the strongly connected components of the
false-positive graph were identified using an iterative
implementation of Tarjan's algorithm, chosen over the recursive
form to accommodate the depth of the graph without exhausting
the Python call stack. Each strongly connected component
corresponds to a group of classes whose mutual false positives
cannot be resolved by reordering. Second, the components were
collapsed to single nodes and the resulting directed acyclic
graph was topologically sorted using Kahn's algorithm; among
components ready for placement at a given step, the component
depended on by the largest number of others was selected first
as a tie-breaker. Third, within each non-trivial component,
classes were ordered by a greedy heuristic in which the node
with the largest difference between incoming and outgoing edge
weights was placed first, with weights updated after each
placement to reflect edges removed from the residual graph.
Classes that participate in no false-positive edge were appended
at the end, where their position has no effect on accuracy.

\subsection{Deterministic ReactionClassifier}\label{sec:methods:classifier}

The ordered SMIRKS database and the LLM-derived hierarchy are combined into a single deployable artefact exposing two inference modes: \emph{Ordered SMIRKS} and \emph{Hybrid strict}. This subsection details the architecture of the gating model, the class--template index, and the inference execution logic.

\subsubsection{Reaction difference fingerprint MLP}
The gating classifier is a multilayer perceptron trained on
reaction difference fingerprints (MDP)
For each reaction we concatenate two folded $2048$-bit fingerprints
computed with the reference Morgan Fingerprint implementation at radius
$r=2$: the product fingerprint, capturing the substructural
context of the product, and the difference fingerprint, encoding
the symmetric difference between reactant and product substructures.
The resulting $4096$-dimensional binary vector is the sole model
input. The MLP has one hidden layer of $512$ GELU units with
dropout~$0.1$, followed by a linear classification head that emits
one logit per class at the deepest available tier of the LLM
hierarchy. The model was trained on the train fold described in
Section~\ref{sec:methods:data} using AdamW (learning rate
$3 \times 10^{-4}$, weight decay $10^{-4}$, batch size $1024$,
$40$ epochs, label smoothing disabled). To prevent leakage between
chemically near-identical reactions, train/validation partitioning
was performed at the level of the retrosynthetic template hash:
reactions sharing a template were placed jointly in the train or
validation fold. Inference returns the argmax class label
$\hat{c}$ and its softmax confidence $p(\hat{c})$; neither value
is treated as a final classification on its own, and we do not
report standalone MDP accuracy because its predictions lack the
determinism guarantee required by the deployment use case.

\subsubsection{Class--template index}
The hybrid classifier requires a lookup from class code to the
templates that compactly represent that class. We use the
\texttt{TEMPLATE\_rr0rp1\_ring0} retrosynthetic templates extracted
during preprocessing (radius~0 on the reactant side, radius~1 on
the product side, ring expansion disabled). For each unique class
code in the LLM-derived hierarchy at the deepest available tier
($6{,}281$ codes at L5), we collect all distinct templates whose
underlying reactions were labelled with that code, yielding a
mapping
\(\texttt{class\_to\_templates}:\ \mathrm{code}\to\{\mathrm{template}_i\}\).
This index is computed once from the working dataset
(Section~\ref{sec:methods:data}) and held in memory for inference.

\subsubsection{Hybrid strict inference}
Given a query reaction $r$, hybrid strict proceeds in three steps. \textbf{(i)~Gate}: the MDP MLP produces a predicted class $\hat{c}$ at the deepest hierarchical tier. \textbf{(ii)~Subset}: the candidate template pool is restricted to the $44{,}449$ exact reaction templates that share a common tier-3 prefix with $\hat{c}$ (the default \texttt{subset\_tier}~$=3$). \textbf{(iii)~Match}: the subsetted templates are applied via RDKit's \texttt{RunReactants}, evaluating all subsets and permutations of the reactant molecules. The predicted label is accepted only when at least one template within the subset fires and produces the correct canonicalised product. If no template fires correctly on the reaction, the strict-match acceptance rule defaults the system to return \texttt{OtherReaction} rather than the MDP prediction, recording an abstention.
\subsubsection{Latency measurement}
Per-reaction latency was measured under 16 parallel workers multiprocessed
inference on a workstation CPU (AMD Ryzen 9 7900X (12 cores / 24 threads), 64 GB RAM).
Each method was timed end-to-end including any model warm-up on
the first query, with the warm-up call excluded from the
distribution. Latency statistics (mean, median, p95) are computed
across the full evaluation set after parallel collection.

\subsection{Evaluation methodology}\label{sec:methods:eval}

\subsubsection{Hierarchical match definitions}
Predictions were compared to ground-truth labels at each tier
(level~1 through level~5) using two complementary criteria. A
\emph{strict match} at tier~$k$ requires the predicted code and
the ground-truth code to agree exactly when both are truncated
to $k$ levels. An \emph{ancestor match} at tier~$k$ accepts the
prediction whenever the predicted code is an ancestor or a
descendant of the ground-truth code at any depth, on the basis
that the prediction lies on the same branch of the hierarchy and
differs only in granularity. The two scores coincide except when
a generic SMIRKS at tier~$k$ is applied to a reaction whose
ground-truth label sits at a deeper tier, or vice versa.

\subsubsection{Coverage categories}
For each evaluated reaction we recorded one of the following
outcomes. \emph{Correct} indicates either a strict match or
agreement after applying the structurally-indistinguishable
class merges described below. \emph{Mismatch} indicates that a
SMIRKS fired but predicted the wrong class. \emph{Missed}
indicates that no SMIRKS in the database fired on the reaction
even though its class is represented in the database.
\emph{Uncovered} indicates that the reaction's class was
excluded from training because it contained fewer than five
reactions; these reactions are retained in the corpus but
excluded from accuracy denominators. \emph{Conflict} indicates
that the ground-truth label carries a \texttt{CONFLICT:} prefix
from the aggregator stage and is therefore ambiguous; these are
also excluded from accuracy denominators.

\subsubsection{Class merges for indistinguishable transformations}
A small number of class pairs in the LLM-derived hierarchy
encode the same molecular transformation but appear under
different parent super-classes due to mechanistic or strategic
context. Buchwald-Hartwig amination, nucleophilic aromatic
substitution, and heteroaryl amination, for example, all proceed
through formation of a C–N bond between the same functional
groups, and ester hydrolysis appears both as a deprotection step
and as a stand-alone functional-group interconversion. We
curated 16 such pairs manually and treated predictions within
each merged group as equivalent during evaluation. The full list
is provided in the Supplementary Information.

\subsection{Software, hardware, and reproducibility}\label{sec:methods:repro}

The classification pipeline was implemented in Python. Inference
was carried out through the Google GenAI SDK; chemical structure
handling and SMIRKS application used RDKit~\citep{rdkit};
atom-to-atom mapping used RXNMapper~\citep{schwaller2021extraction};
and label-noise estimation used the
\texttt{cleanlab}~\citep{northcutt2021confident} package.
Reaction and molecular fingerprints were computed with the
reference \texttt{rdkit} implementation~\citep{rdkit}.
Generalisation, false-positive testing, and ordering used
\texttt{joblib} for parallelism and \texttt{Pydantic} for
structured LLM output, with the strongly-connected-component and
topological-sort routines implemented directly. A fixed random
seed of $210{,}995$ was used wherever stochastic behaviour is
involved (cross-validation folds, train/test splits). Inference
parallelism was set to eight worker processes or threads
throughout.

\begin{ack}

\textbf{Acknowledgements.}
The authors thank collaborators in EPFL Laboratory of Artificial Chemical Intelligence (LIAC) for helpful discussions

\textbf{Funding.} This work was supported by the Swiss National Science Foundation through the National Centre of Competence in Research (NCCR) Catalysis (225147) and through the grant (214915). Maarten Dobbelaere acknowledges financial support from the Research Foundation -- Flanders (FWO Vlaanderen) through postdoctoral fellowship grant 1266226N and travel grant V414426N. In addition this research was conducted with support from Google.org and the Google Cloud Research Credits program for the Gemini Academic Program.

\textbf{Competing interests.} The authors declare no competing interests.

\textbf{Code availability.} Deterministic ReactionClassifier is released under an open-source licence
  at the project repository. \url{https://github.com/schwallergroup/ReactionClassifier.git}

\end{ack}

\bibliographystyle{naturemag}
\bibliography{sn-bibliography}


\appendix

\title{Supporting Information}
\begin{center}
{\large S1: Label Diversity}
\end{center}

\section{Classification Summary Statistics}
\label{sec:summary}

A total of 179,495 unique reactions across 42,125 retrosynthetic templates were classified
into 12,650 distinct class codes, drawn from a label vocabulary of 14,073 hierarchical class
labels (plus 9,654 condition labels). By assigning each template its majority class and
extrapolating to the full template occurrence count, the classification covers an estimated
860,675 reactions. The hierarchy spans 7 levels: 19 super-classes (L1), 106 sub-classes (L2),
1,546 types (L3), 6,224 subtypes (L4), and progressively finer distinctions at L5--L7.
Within templates, classification is highly consistent: 99.97\% of templates (42,112/42,125)
are assigned a single unique class code across all their classified reactions.

\subsection{Class-size distributions}

\begin{figure}[htbp]
\centering
\includegraphics[width=\textwidth]{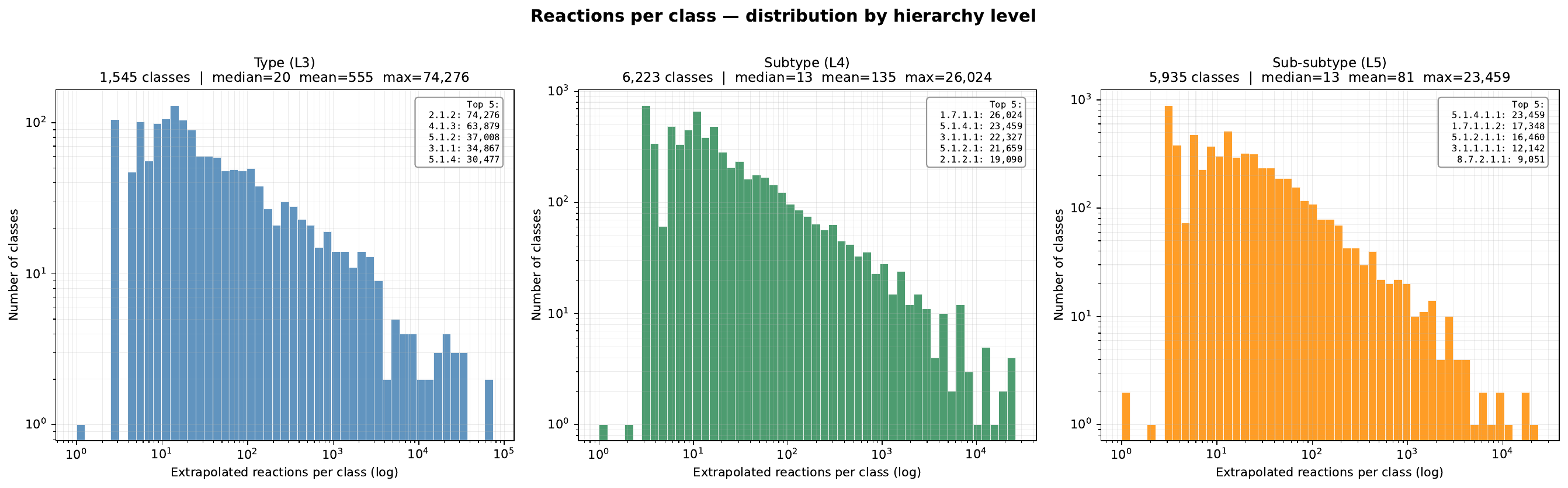}
\caption{Distribution of extrapolated reactions per class at three hierarchy levels.
  L3 (Type): 1,545 classes, median 20, max 74,276.
  L4 (Subtype): 6,223 classes, median 13, max 26,024.
  L5 (Sub-subtype): 5,935 classes, median 13, max 23,459.}
\label{fig:distributions}
\end{figure}

Figure~\ref{fig:distributions} shows the distribution of extrapolated reaction counts per
class at three hierarchy levels (L3, L4, L5). All three exhibit a power-law shape: the
majority of classes contain fewer than 50 reactions, while a long tail of high-frequency
classes dominates the total reaction count.

\subsection{Label diversity across L3 classes}

\begin{figure}[htbp]
\centering
\includegraphics[width=\textwidth]{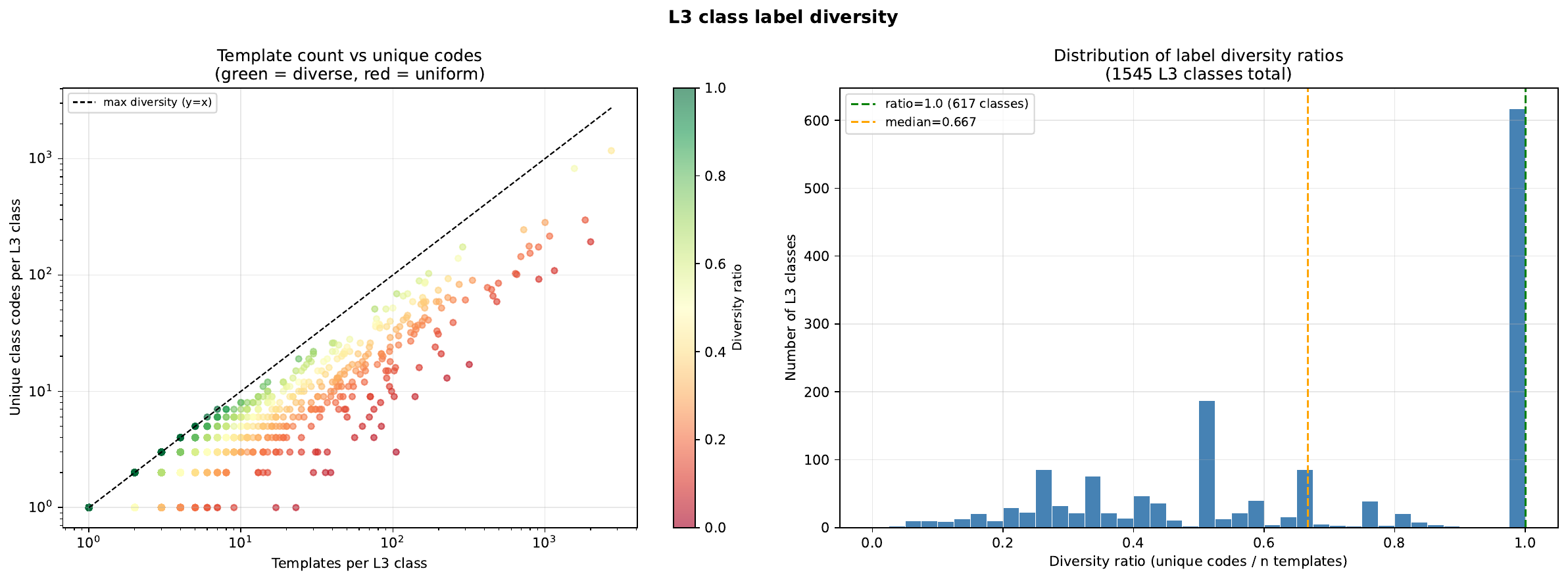}
\caption{Label diversity across L3 classes. \textbf{Left:} scatter plot of template count
  vs.\ unique class codes per L3 class, coloured by diversity ratio (green = diverse,
  red = uniform). The dashed line indicates maximal diversity ($y = x$). \textbf{Right:}
  histogram of diversity ratios across all 1,545 L3 classes.}
\label{fig:diversity}
\end{figure}

To quantify how uniformly the LLM assigns fine-grained subtypes within a given reaction type,
we define the \emph{label diversity ratio} for each L3 class as the number of unique
finest-level class codes divided by the number of templates. A ratio of 1.0 indicates maximal
diversity (every template received a unique subtype), while low ratios indicate uniform
assignment. Figure~\ref{fig:diversity} shows the distribution: the median diversity ratio is
0.667, 617 L3 classes achieve maximal diversity (ratio~=~1.0), and only 22 fall below 0.1.

\clearpage
\section{Maximally Diverse L3 Classes: Worked Examples}
\label{sec:examples}

We present two L3 classes with maximal label diversity (ratio = 1.0) as illustrative
examples. In both cases, every classified template maps to a unique subtype code, reflecting
genuine chemical variation within the reaction family.

\subsection{3.10.3 Aromatic Formylation}
\label{sec:formylation}

This L3 class contains 7 classified templates, each assigned a unique subtype code.
All reactions install an aldehyde (--CHO) onto an aromatic ring, but the reagent, mechanism,
and substrate scope differ for every template.

\begin{table}[htbp]
\centering
\caption{Aromatic Formylation subtypes (3.10.3). Each template maps to a unique class code.}
\label{tab:formylation}
\small
\begin{tabular}{@{}llp{5.5cm}@{}}
\toprule
Code & Method & Key reagents \\
\midrule
3.10.3.2.2.1 & Vilsmeier--Haack             & DMF, POCl\textsubscript{3} \\
3.10.3.4     & Lithiation--formylation       & \textit{n}-BuLi, DMF \\
3.10.3.5     & Rieche (heterocycle, pyrrole) & MOM-Cl, AlCl\textsubscript{3} \\
3.10.3.5.1   & Rieche (heterocycle, thiophene)& MOM-Cl, AlCl\textsubscript{3} \\
3.10.3.6     & Rieche (arene)                & MOM-Cl, TiCl\textsubscript{4} \\
3.10.3.6.1   & Duff-type (ester formyl source)& EtOAc, Lewis acid \\
3.10.3.7     & Vilsmeier--Haack (toluene)    & Toluene, DMF, POCl\textsubscript{3} \\
\bottomrule
\end{tabular}
\end{table}

\begin{figure}[htbp]
\centering
\begin{subfigure}[b]{\textwidth}
  \centering
  \includegraphics[width=0.85\textwidth]{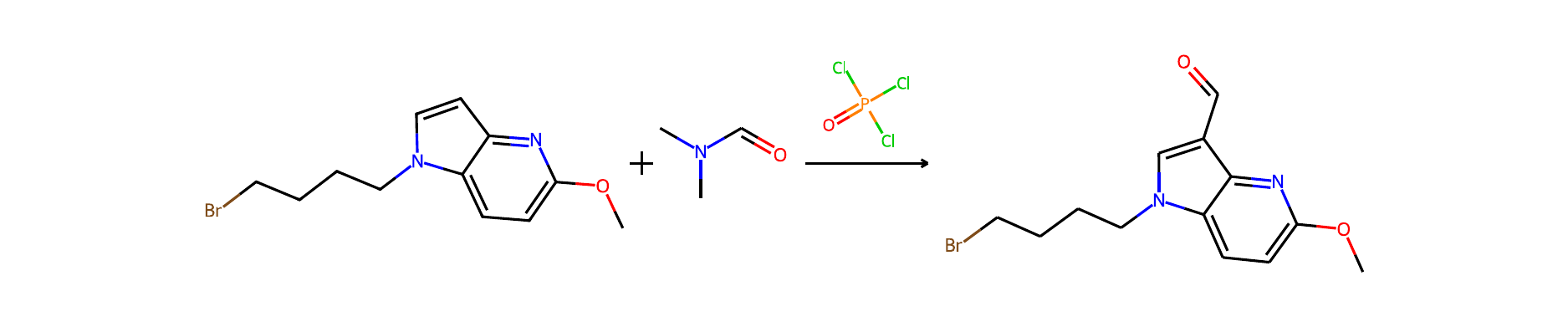}
  \caption{\textbf{3.10.3.2.2.1} Vilsmeier--Haack formylation of a pyrrolo[1,2-\textit{a}]pyrimidine with DMF/POCl\textsubscript{3}.}
\end{subfigure}

\vspace{0.4cm}
\begin{subfigure}[b]{\textwidth}
  \centering
  \includegraphics[width=0.85\textwidth]{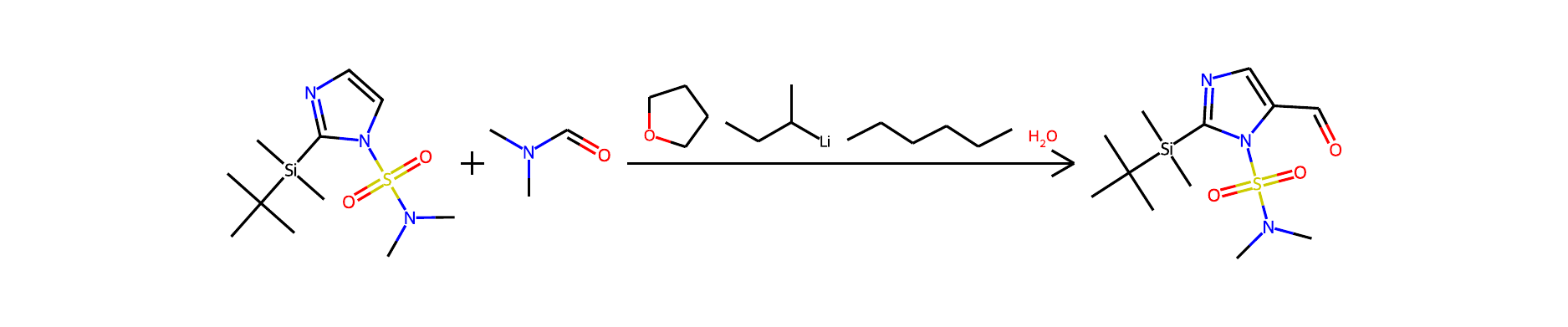}
  \caption{\textbf{3.10.3.4} Lithiation--formylation of an imidazole with \textit{n}-BuLi then DMF.}
\end{subfigure}

\vspace{0.4cm}
\begin{subfigure}[b]{\textwidth}
  \centering
  \includegraphics[width=0.85\textwidth]{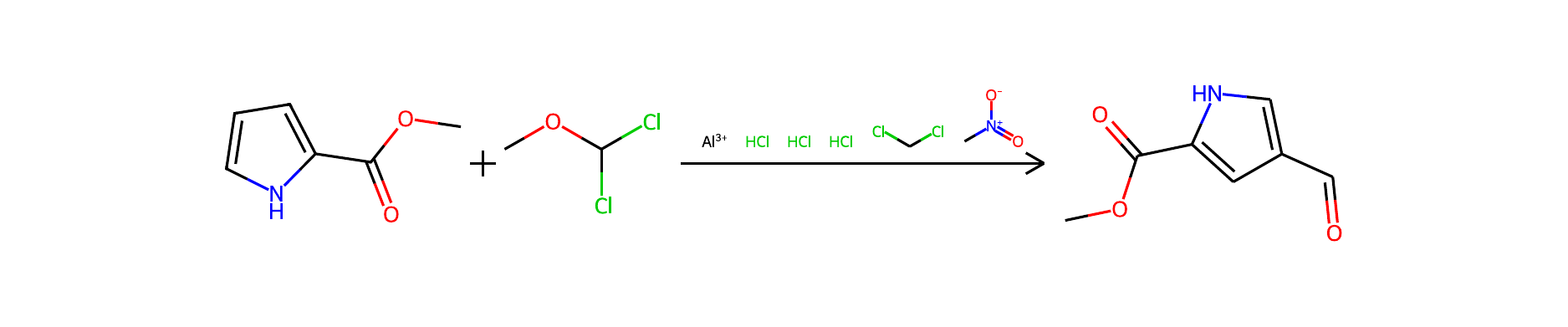}
  \caption{\textbf{3.10.3.5} Rieche formylation of a pyrrole ester with MOM-Cl/AlCl\textsubscript{3}.}
\end{subfigure}

\vspace{0.4cm}
\begin{subfigure}[b]{\textwidth}
  \centering
  \includegraphics[width=0.85\textwidth]{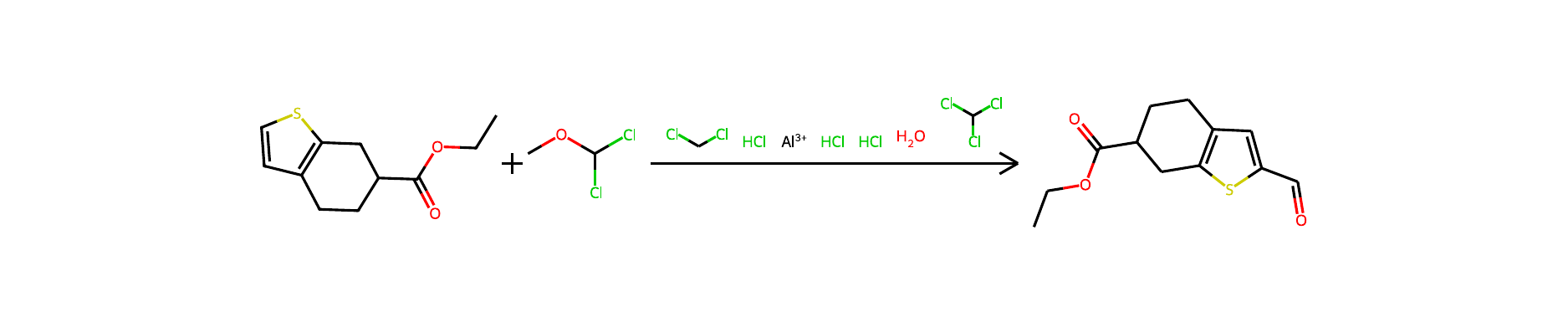}
  \caption{\textbf{3.10.3.5.1} Rieche formylation of a thiophene with MOM-Cl/AlCl\textsubscript{3}.}
\end{subfigure}
\end{figure}
\begin{figure}[htbp]
\vspace{0.4cm}
\begin{subfigure}[b]{\textwidth}
  \centering
  \includegraphics[width=0.85\textwidth]{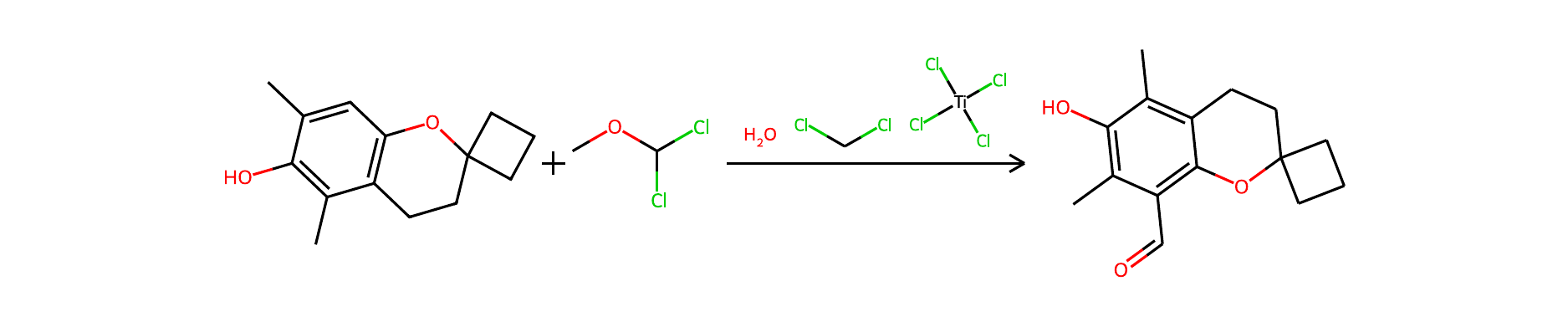}
  \caption{\textbf{3.10.3.6} Rieche formylation of a phenol with MOM-Cl/TiCl\textsubscript{4}.}
\end{subfigure}

\centering
\vspace{0.4cm}
\begin{subfigure}[b]{\textwidth}
  \centering
  \includegraphics[width=0.85\textwidth]{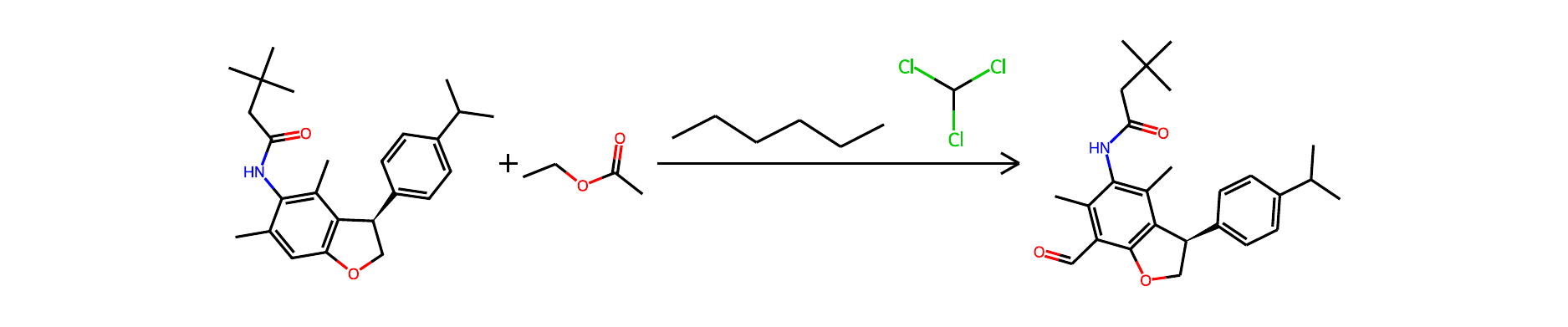}
  \caption{\textbf{3.10.3.6.1} Duff-type formylation using ethyl acetate as formyl source.}
\end{subfigure}

\vspace{0.4cm}
\begin{subfigure}[b]{\textwidth}
  \centering
  \includegraphics[width=0.85\textwidth]{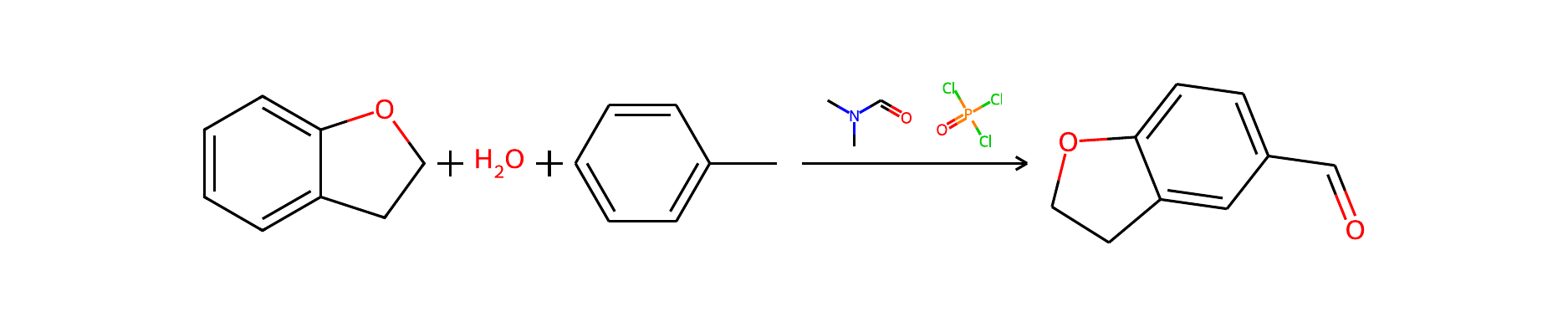}
  \caption{\textbf{3.10.3.7} Vilsmeier--Haack formylation using toluene, DMF, and POCl\textsubscript{3}.}
\end{subfigure}

\caption{Representative reactions for each subtype in class 3.10.3 (Aromatic Formylation).}
\label{fig:formylation}
\end{figure}

\clearpage
\subsection{9.1.1 Hydrohalogenation of Alkenes}
\label{sec:hydrohal}

This L3 class contains 6 templates, each with a unique code.
All reactions add halogen(s) across a C=C double bond, but they span Markovnikov \textit{vs.}\
anti-Markovnikov selectivity, different halogens (Br, F), radical \textit{vs.}\ ionic mechanisms,
and multi-component additions (azidohalogenation).

\begin{table}[htbp]
\centering
\caption{Hydrohalogenation subtypes (9.1.1). Each template maps to a unique class code.}
\label{tab:hydrohal}
\small
\begin{tabular}{@{}lp{6.5cm}@{}}
\toprule
Code & Variant \\
\midrule
9.1.1.3.1.2 & Markovnikov HBr addition (benzylic position) \\
9.1.1.3.2   & Anti-Markovnikov HBr addition (radical, terminal alkene) \\
9.1.1.3.3   & HF addition to perfluoroalkene \\
9.1.1.1.2   & Bromofluorination (NBS + HF $\to$ vicinal BrF) \\
9.1.1.6     & Br\textsubscript{2} radical addition (BPO initiator) \\
9.1.1.7.1   & Azidohalogenation (NBS + NaN\textsubscript{3}) \\
\bottomrule
\end{tabular}
\end{table}

\begin{figure}[htbp]
\centering
\begin{subfigure}[b]{\textwidth}
  \centering
  \includegraphics[width=0.85\textwidth]{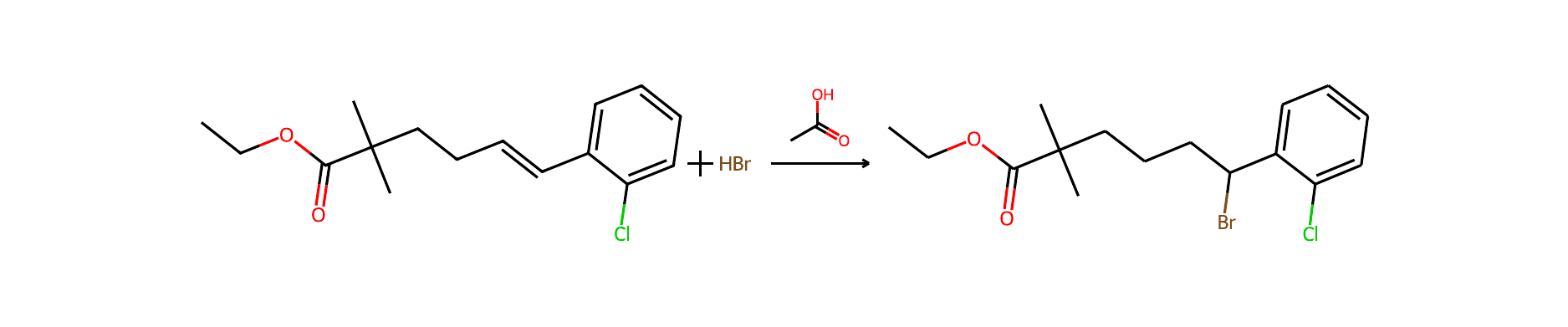}
  \caption{\textbf{9.1.1.3.1.2} Markovnikov HBr addition to a styrenyl alkene (benzylic bromide product).}
\end{subfigure}

\vspace{0.4cm}
\begin{subfigure}[b]{\textwidth}
  \centering
  \includegraphics[width=0.85\textwidth]{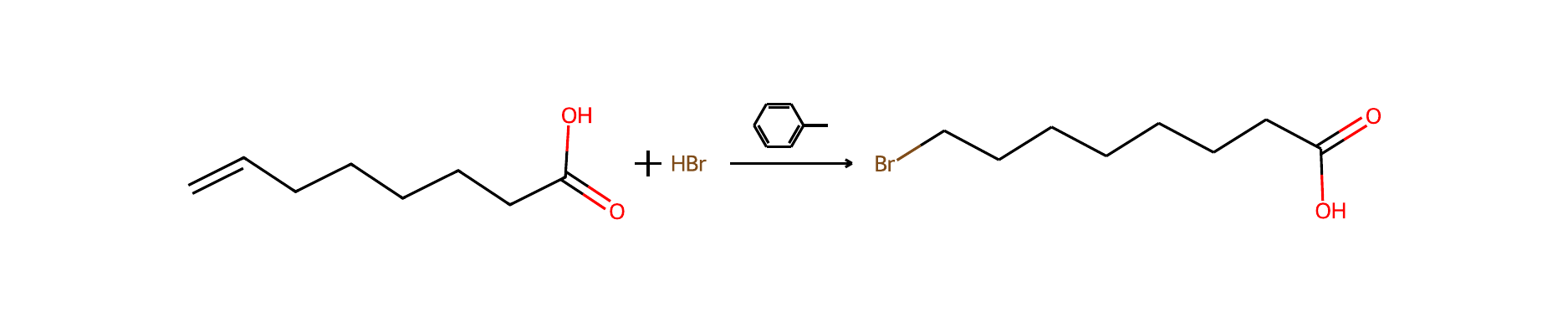}
  \caption{\textbf{9.1.1.3.2} Anti-Markovnikov HBr addition to a terminal alkene (toluene solvent, radical conditions).}
\end{subfigure}

\vspace{0.4cm}
\begin{subfigure}[b]{\textwidth}
  \centering
  \includegraphics[width=0.85\textwidth]{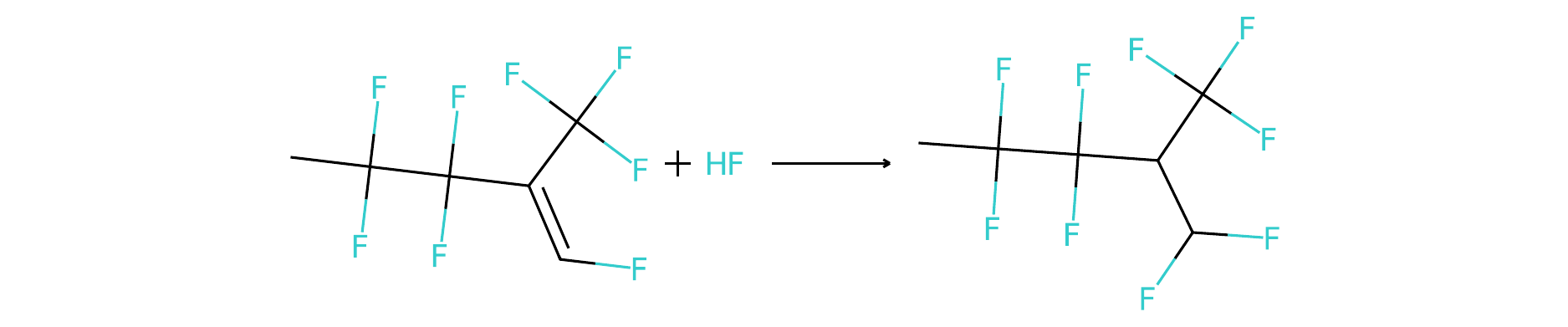}
  \caption{\textbf{9.1.1.3.3} HF addition across a perfluoroalkene.}
\end{subfigure}

\vspace{0.4cm}
\begin{subfigure}[b]{\textwidth}
  \centering
  \includegraphics[width=0.85\textwidth]{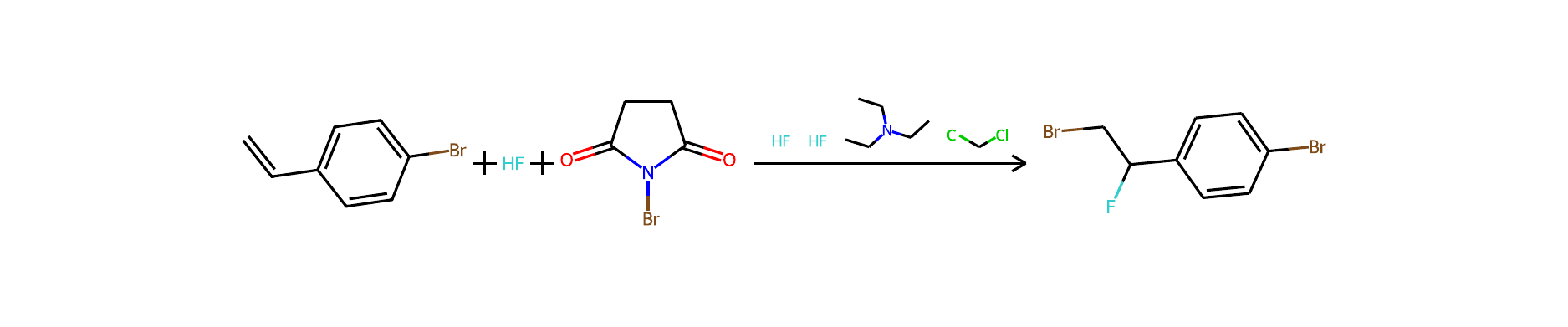}
  \caption{\textbf{9.1.1.1.2} Bromofluorination: NBS + HF gives vicinal bromo-fluoro product.}
\end{subfigure}

\vspace{0.4cm}
\begin{subfigure}[b]{\textwidth}
  \centering
  \includegraphics[width=0.85\textwidth]{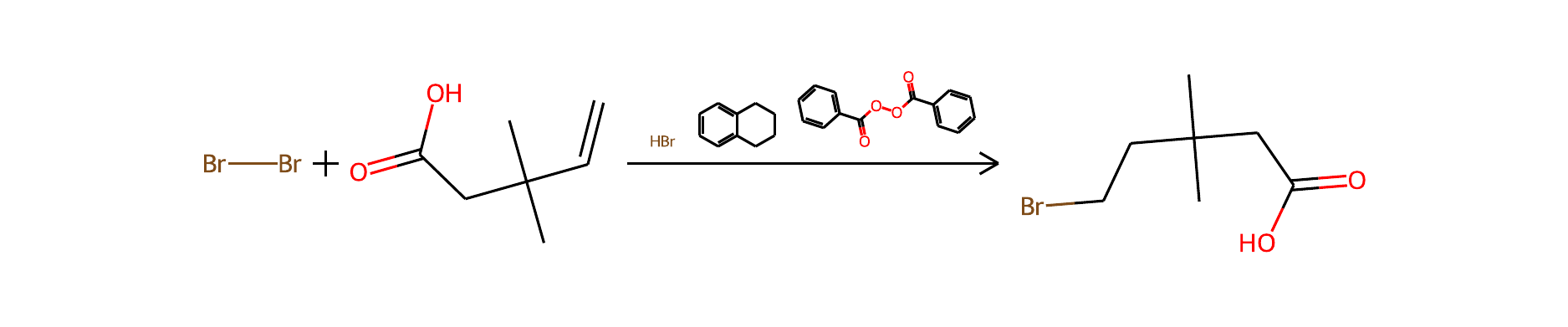}
  \caption{\textbf{9.1.1.6} Radical Br\textsubscript{2} addition with benzoyl peroxide (BPO) initiator; anti-Markovnikov terminal bromide.}
\end{subfigure}

\vspace{0.4cm}
\begin{subfigure}[b]{\textwidth}
  \centering
  \includegraphics[width=0.85\textwidth]{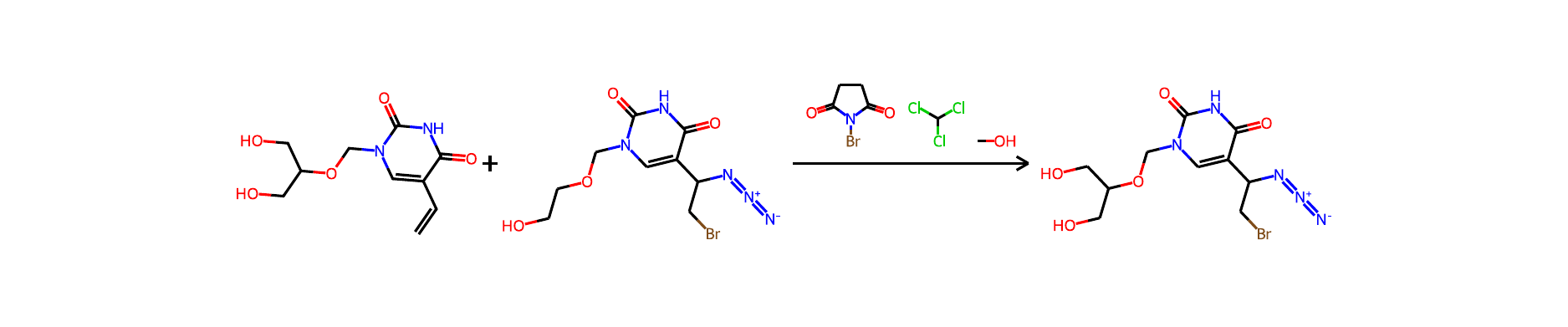}
  \caption{\textbf{9.1.1.7.1} Azidohalogenation: NBS introduces Br while azide adds to give vicinal azido-bromide on a uracil substrate.}
\end{subfigure}

\caption{Representative reactions for each subtype in class 9.1.1 (Hydrohalogenation of Alkenes).}
\label{fig:hydrohal}
\end{figure}
\section{Classification Pipeline}
\label{sec:pipeline}

\subsection{Overview}

We developed a five-stage agentic classification pipeline in which each stage is a
specialised large language model (LLM) call (Gemini~2.5 Flash or Pro via Google Vertex~AI,
temperature 0.1--0.2).  The pipeline takes as input a set of atom-mapped reaction SMILES,
each associated with a retrosynthetic template hash and a SMARTS-encoded retro-template,
and produces a hierarchical class code together with human-readable names at every level
of the taxonomy.  A central design principle is the \emph{two-phase context strategy}:
a lightweight first stage classifies every reaction against the full (but compact) two-level
hierarchy (\(\sim\)100 entries), after which all subsequent stages operate on only the
relevant \emph{subtree} of the full taxonomy.  This is necessary because the complete
taxonomy, which grows to over 14\,000 entries during a run, far exceeds the context
window that an LLM can use effectively; by routing first to a two-level parent code, the
detailed agent receives a focused subtree of at most a few hundred entries, enabling precise
fine-grained classification without dilution from irrelevant branches.

\subsection{Template-level cohort batching}

Reactions are grouped into \emph{cohorts} by retrosynthetic template hash before
entering the pipeline.  All reactions that share the same template undergo the same core
bond disconnection and therefore, by construction, represent the same mechanistic
transformation regardless of peripheral substituents.  Each cohort contains up to 5
representative reactions; the pipeline classifies the cohort once and broadcasts the
result to all members.  This methodology we hypothesise acts as a "batch regularisation" and increases the likelyhood of a correct classification.

\subsection{Stage 1: Hierarchy classification}

The hierarchy agent receives the full two-level reaction taxonomy (9 super-classes,
52 sub-classes) as part of its system prompt, together
with the reaction SMILES (optionally with atom-mapping numbers stripped via RDKit to
reduce input complexity).  The agent is instructed to identify the key bonds formed or
broken and the functional groups involved, and to classify the reaction into the most
appropriate sub-class using a \emph{reagent-agnostic} analysis: the core structural
transformation must be identifiable without reference to catalysts or reagents.  The
output is a single two-level code (e.g., \texttt{2.1}) per cohort, returned in XML
format (\texttt{<reaction\_class>X.Y</reaction\_class>}).

\subsection{Stage 2: Detailed classification}

The detailed agent receives only the \emph{subtree} of the hierarchy rooted at the
two-level parent code assigned in Stage~1.  This subtree is typically 10--200
entries and contains all known types (L3), subtypes (L4), sub-subtypes (L5), and
condition variants (L6+) beneath the parent.  The agent follows a strict two-path
decision algorithm:

\begin{description}
  \item[Path A (match found):] Step~1 identifies the Level~3/4 branch by examining
    \emph{only} reactant and product functional groups (reagent-agnostic). Step~2 then
    examines reagents and catalysts to select a Level~5+ condition variant within the
    already-locked branch.
  \item[Path B (no match):] If no existing Level~3/4 entry matches the core
    transformation, the agent terminates with an \texttt{.Other} suffix appended to
    the parent code, signalling that a new entry is needed.
\end{description}

\noindent The reaction is presented with explicit role labels
(\texttt{reactants:}, \texttt{reagents:}, \texttt{product:}) together with the
retrosynthetic SMARTS template.  The output consists of five XML tags:
\texttt{<reaction\_super\_class>}, \texttt{<reaction\_sub\_class>},
\texttt{<reaction\_type>}, \texttt{<reaction\_subtype>}, and
\texttt{<class\_number>}.

\subsection{Stage 3: Verification}

A verification agent audits each detailed classification.  It performs two checks in
sequence:

\begin{enumerate}
  \item \textbf{Hierarchy integrity} (primary check): Does the reaction fundamentally
    belong in this super-class and sub-class?  Fatal errors such as a reduction
    classified under oxidation, or a C--C bond formation under protection/deprotection trigger
    an \texttt{incorrect\_hierarchy} flag, and the cohort is excluded from further
    processing entirely.
  \item \textbf{Functional specificity} (secondary check): Do the named functional
    groups in the class label precisely match the reactant structures?  Do the Level~5+
    condition labels match the reagents present?  If not, the classification is marked
    as a mismatch (\texttt{match=false}) with a one-sentence reason, and the cohort is
    routed to the generator stage.
\end{enumerate}

\noindent Cohorts that pass both checks (\texttt{match=true}) exit the pipeline with
their Stage~2 classification as final.

\subsection{Stage 4: Proposal generation}

For cohorts flagged as mismatches, a generator agent proposes new taxonomy entries.
The agent receives the same hierarchy subtree and reaction data as the detailed agent,
plus the failed classification and the verifier's reason for rejection.  It follows a
``What vs.\ How'' principle:

\begin{itemize}
  \item \textbf{Levels 1--4 (``What''):} Describe the reagent-agnostic core
    transformation by specifying the reactant functional groups and the bond(s) formed.
  \item \textbf{Levels 5+ (``How''):} Describe the specific reagent system, named
    reaction, or condition set (e.g., ``cond: EDC/HOBt Coupling'').
\end{itemize}

\noindent The output includes a suggested parent code, a new entry code and name,
the exact hierarchy line to append, and a rationale.  The agent may also propose
intermediate parent entries if they are missing from the current subtree.

\subsection{Stage 5: Aggregation and dynamic mapping}

The aggregator agent receives all proposals for a given two-level parent code (chunked
into groups of up to 50 cohorts) and consolidates them into a unified set of new
hierarchy entries plus a mapping from each cohort to its final class code.  The
aggregator deduplicates overlapping proposals, for example ``Acid chloride + primary amine to secondary amide'' entry are merged into a single
canonical Level~3/4 class, with separate Level~5+ condition variants if their reagent
systems differ.

The resulting new hierarchy lines are applied to the \emph{dynamic mapping}; an
in-memory dictionary keyed by two-level parent code, where each value is a sorted list
of descendant entries.  This update happens \emph{immediately}, so that subsequent
aggregator calls within the same pipeline run see the newly created entries and can
assign cohorts to them rather than creating duplicates.  The mapping is also
persisted to disk via an atomic write (temporary file $\to$ \texttt{fsync} $\to$
rename) to ensure crash safety.

\subsubsection{Conflict detection and resolution}

A global code-to-label lookup table tracks every class code assigned during the run.
When a new hierarchy line is proposed whose code already exists but with a
\emph{different label}, the line is rejected and recorded as a conflict.  Any cohort
whose final assignment points to a conflicting code receives a
\texttt{CONFLICT:X.Y.Z.W} prefix in the output \texttt{LLM class number} column,
with all text label columns left empty.  This makes conflicts trivially filterable
in downstream analysis and flags them for manual resolution.  Conflicts are
intentionally \emph{not} retried automatically: because the same code with two
different labels indicates genuine taxonomic ambiguity, automated resolution risks
silently merging chemically distinct categories.

\subsubsection{Hierarchy growth}

The taxonomy is not static.  Over the course of a classification run, the
verify$\to$generate$\to$aggregate loop extends the hierarchy with new entries proposed
by the LLM and validated by the aggregator.  In the present dataset, the label
vocabulary grew from an initial seed of $\sim$2\,000 entries to 14\,073 class labels
and 9\,654 condition annotations (23\,727 total), spanning seven hierarchy levels.
This closed-loop design enables the system to handle novel chemistry that falls outside
the initial taxonomy without human intervention, while the conflict detection mechanism
ensures that growth remains consistent.

\subsection{Output format}

The pipeline produces a CSV file with one row per input reaction and the following
classification columns: \texttt{LLM reaction super class} (L1 name),
\texttt{LLM reaction sub class} (L2 name), \texttt{LLM reaction type} (L3 name),
\texttt{LLM reaction subtype} (L4+ name), \texttt{LLM class number} (full numeric
code, e.g., \texttt{2.1.1.3}), and \texttt{classification} (type-level label, e.g.,
\texttt{2.1.1 Amide Schotten-Baumann}).  A companion dynamic mapping JSON file
records the final state of the taxonomy, keyed by two-level parent code.

\section{LLM label noise analysis}
\label{si:label_noise}

We applied the confident learning framework (\textsc{cleanlab}; \citealt{northcutt2021confident}) to estimate label noise for both the LLM (level-3) and NameRXN annotations independently. A two-layer MLP (512 hidden units, SELU activation, dropout 0.1) was trained on concatenated difference and product DRFP fingerprints with five-fold stratified cross-validation; out-of-fold softmax probabilities were used to rank suspected mislabels by self-confidence. On the training set, cleanlab flagged 2.19\% of LLM labels and 0.59\% of NameRXN labels as likely incorrect (Table~\ref{tab:label_noise}).

To distinguish genuine errors from taxonomy-boundary ambiguity in the LLM labels, we inspected all 321 confusion pairs among the 600 reactions flagged on the held-out test set ($n = 30{,}802$). We classified each pair into one of three categories: (i)~\emph{boundary ambiguity}, where both the given and predicted labels describe the same underlying transformation and differ only in taxonomic granularity or naming convention; (ii)~\emph{partial overlap}, where the two classes share a common mechanistic step but differ in scope; and (iii)~\emph{genuine error}, where the given and predicted classes are mechanistically distinct. Representative examples of each category are listed in Supplementary Table~\ref{tab:llm_confusion_pairs}.

Of the 600 flagged reactions, 487 (81.2\%) fell into category~(i), 11 LLM confusion pairs into category~(ii), and 102 (17.0\%) into category~(iii). The most frequent boundary-ambiguity pair was \texttt{1.3.6} \emph{Amination of Heteroaryl Halides} $\leftrightarrow$ \texttt{1.3.5} \emph{Nucleophilic Aromatic Substitution} (117 reactions combined), both of which describe displacement of a leaving group on an electron-poor aromatic ring by a nitrogen nucleophile but differ in whether the taxonomy privileges substrate identity or mechanism. Other recurrent boundary pairs included \texttt{5.1.9} \emph{Cleavage of Phenol Protecting Groups} $\to$ \texttt{5.1.1} \emph{Cleavage of Alcohol Protecting Groups} (14 reactions), reflecting phenols being a subset of alcohols. Category~(iii) errors were diverse and included reductions mislabelled as oxidations and \ch{SNAr} reactions confused with $N$-alkylation. After restricting to category~(iii) only, the adjusted mislabelling rate is 0.33\%, comparable to the 0.59\% noise estimated for NameRXN by the same method.

\begin{longtable}{r l c >{\raggedright\arraybackslash}p{5cm} c >{\raggedright\arraybackslash}p{5cm}}
\caption{LLM level-3 label confusion pairs identified by cleanlab on the
training set (n = 538{,}029). \emph{Verdict}: \textbf{FAIR} = taxonomy
boundary ambiguity (classes are genuinely similar); \textbf{SUSPECT} =
reaction types are mechanistically distinct, indicating a probable labelling
error; \textbf{UNCLEAR} = one class code unmapped.}
\label{tab:llm_confusion_pairs} \\
\toprule
\# & $n$ & Verdict & Given label & $\to$ & Model prediction \\
\midrule
\endfirsthead
\multicolumn{6}{c}{\tablename\ \thetable{} -- \textit{continued}} \\
\toprule
\# & $n$ & Verdict & Given label & $\to$ & Model prediction \\
\midrule
\endhead
\midrule
\multicolumn{6}{r}{\textit{Continued on next page}} \\
\endfoot
\bottomrule
\endlastfoot
  1 & 72 & \textcolor{teal}{\textbf{FAIR}} & \texttt{1.3.6} Amination of Heteroaryl Halides & & \texttt{1.3.5} SNAr$^\dagger$ \\
  2 & 45 & \textcolor{teal}{\textbf{FAIR}} & \texttt{1.3.5} SNAr & & \texttt{1.3.6} Amination of Heteroaryl Halides$^\dagger$ \\
  3 & 14 & \textcolor{teal}{\textbf{FAIR}} & \texttt{5.1.9} Cleavage of Phenol PGs & & \texttt{5.1.1} Cleavage of Alcohol PGs$^\dagger$ \\
  4 & 12 & \textcolor{teal}{\textbf{FAIR}} & \texttt{6.1.16} Reduction of nitrobenzylamines & & \texttt{6.1.11} Reduction of nitrobenzenes to anilines$^\dagger$ \\
  5 & 12 & \textcolor{teal}{\textbf{FAIR}} & \texttt{6.5.5} Reduction of aryl ketones & & \texttt{6.5.1} Hydride Reduction$^\dagger$ \\
  6 & 9 & \textcolor{teal}{\textbf{FAIR}} & \texttt{6.1} NO$_2$ to amine & & \texttt{6.1.11} Reduction of nitrobenzenes to anilines \\
  7 & 8 & \textcolor{teal}{\textbf{FAIR}} & \texttt{1.6.5} N-Alkylation of Pyridones/Uracils & & \texttt{1.6.1} N-Alkylation of $\pi$-Excessive Heterocycles$^\dagger$ \\
  8 & 7 & \textcolor{teal}{\textbf{FAIR}} & \texttt{1.7.7} O-Alkylation for PG Formation & & \texttt{1.7.1} Williamson Ether Synthesis$^\dagger$ \\
  9 & 6 & \textcolor{teal}{\textbf{FAIR}} & \texttt{5.1.4} Cleavage of COOH PGs & & \texttt{8.7.2} Carboxylic Acid Derivative Interconversions \\
  10 & 6 & \textcolor{teal}{\textbf{FAIR}} & \texttt{8.7.7} Synthesis of Organoboron Compounds & & \texttt{3.5.8} Pd-Catalyzed Borylation \\
  11 & 5 & \textcolor{teal}{\textbf{FAIR}} & \texttt{1.1.1} Alkylation & & \texttt{1.5.1} N-Alkylation with alkyl halides \\
  12 & 5 & \textcolor{teal}{\textbf{FAIR}} & \texttt{7.1.2} Oxidation of Secondary Alcohols to Ketones & & \texttt{7.8.7} Oxidation of Alcohols \\
  13 & 5 & \textcolor{teal}{\textbf{FAIR}} & \texttt{7.8.7} Oxidation of Alcohols & & \texttt{7.1.2} Oxidation of Secondary Alcohols to Ketones \\
  14 & 4 & \textcolor{teal}{\textbf{FAIR}} & \texttt{1.5.1} N-Alkylation with alkyl halides & & \texttt{1.1.1} Alkylation SN2 \\
  15 & 4 & \textcolor{teal}{\textbf{FAIR}} & \texttt{3.11.24} Condensation of ketones with amide acetals & & \texttt{1.1.9} Condensation with Amide Acetals \\
  16 & 4 & \textcolor{teal}{\textbf{FAIR}} & \texttt{3.3.5} Coupling of Heteroaryl Halides & & \texttt{3.3.1} Coupling of Aryl/Vinyl Halides$^\dagger$ \\
  17 & 4 & \textcolor{teal}{\textbf{FAIR}} & \texttt{6.1.7} Reduction of nitrobenzenesulfonamides & & \texttt{6.1.11} Reduction of nitrobenzenes$^\dagger$ \\
  18 & 3 & \textcolor{teal}{\textbf{FAIR}} & \texttt{1.3.1} Buchwald-Hartwig & & \texttt{1.3.5} SNAr$^\dagger$ \\
  19 & 3 & \textcolor{teal}{\textbf{FAIR}} & \texttt{1.3.1} Buchwald-Hartwig & & \texttt{1.3.6} Amination of Heteroaryl Halides$^\dagger$ \\
  20 & 3 & \textcolor{teal}{\textbf{FAIR}} & \texttt{1.3.2} Ullmann & & \texttt{1.3.8} Amination of Aryl Halides$^\dagger$ \\
  21 & 3 & \textcolor{teal}{\textbf{FAIR}} & \texttt{1.3.6} Amination of Heteroaryl Halides & & \texttt{1.3.1} Buchwald-Hartwig$^\dagger$ \\
  22 & 3 & \textcolor{teal}{\textbf{FAIR}} & \texttt{1.3.8} Amination of Aryl Halides & & \texttt{1.3.1} Buchwald-Hartwig$^\dagger$ \\
  23 & 3 & \textcolor{teal}{\textbf{FAIR}} & \texttt{1.4.2} N-Alkylation of Lactams & & \texttt{1.4.1} N-Alkylation of Amides$^\dagger$ \\
  24 & 3 & \textcolor{teal}{\textbf{FAIR}} & \texttt{1.6.33} N-Alkylation with Benzyl Halides & & \texttt{1.6.1} N-Alkylation of $\pi$-Excessive Heterocycles$^\dagger$ \\
  25 & 3 & \textcolor{teal}{\textbf{FAIR}} & \texttt{1.7.1} Williamson Ether & & \texttt{1.7.7} O-Alkylation for PG Formation$^\dagger$ \\
  26 & 3 & \textcolor{gray}{\textbf{UNCLEAR}} & \texttt{3.10.1} Friedel-Crafts Acylation & & \texttt{3.Other} ??? \\
  27 & 3 & \textcolor{teal}{\textbf{FAIR}} & \texttt{3.11.10} Cyanation of aryl halides & & \texttt{3.5.7} Pd-Catalyzed Cyanation \\
  28 & 3 & \textcolor{teal}{\textbf{FAIR}} & \texttt{3.11.69} Borylation of aryl/heteroaryl halides & & \texttt{3.5.8} Pd-Catalyzed Borylation \\
  29 & 3 & \textcolor{teal}{\textbf{FAIR}} & \texttt{3.2} Heck Reaction & & \texttt{3.2.1} Standard Heck reaction \\
  30 & 3 & \textcolor{red}{\textbf{SUSPECT}} & \texttt{6.2.4} Condensation of oxindoles & & \texttt{3.11.5} Aromatic Aldehyde Condensations \\
  31 & 3 & \textcolor{teal}{\textbf{FAIR}} & \texttt{6.5.1} Hydride Reduction & & \texttt{6.5.5} Reduction of aryl ketones$^\dagger$ \\
  32 & 3 & \textcolor{teal}{\textbf{FAIR}} & \texttt{6.6.1} Catalytic Hydrogenation & & \texttt{6.6.4} Reduction of styrenes$^\dagger$ \\
  33 & 3 & \textcolor{teal}{\textbf{FAIR}} & \texttt{6.6.4} Reduction of styrenes & & \texttt{6.6.1} Catalytic Hydrogenation$^\dagger$ \\
  34 & 3 & \textcolor{red}{\textbf{SUSPECT}} & \texttt{6.9.2} Reductive Cleavage and Decarboxylation & & \texttt{1.3.6} Amination of Heteroaryl Halides \\
  35 & 3 & \textcolor{red}{\textbf{SUSPECT}} & \texttt{9.1.4} Halogenation of Heteroarenes & & \texttt{8.7.4} Amine Derivatizations \\
  36 & 3 & \textcolor{teal}{\textbf{FAIR}} & \texttt{9.1.4} Halogenation of Heteroarenes & & \texttt{9.1.5} Halogenation of Arenes$^\dagger$ \\
  37 & 2 & \textcolor{teal}{\textbf{FAIR}} & \texttt{1.1.1} SN2 & & \texttt{1.2.8} N-alkylation with benzyl halides \\
  38 & 2 & \textcolor{teal}{\textbf{FAIR}} & \texttt{1.1.1} SN2 & & \texttt{1.4.3} N-Alkylation of Imides \\
  39 & 2 & \textcolor{teal}{\textbf{FAIR}} & \texttt{1.1.9} Condensation with Amide Acetals & & \texttt{3.11.24} Condensation of ketones with amide acetals \\
  40 & 2 & \textcolor{teal}{\textbf{FAIR}} & \texttt{1.2.12} Chlorination of heteroaryl hydroxy compounds & & \texttt{8.1.5} Heteroaromatic Alcohol to Heteroaryl Chloride \\
  41 & 2 & \textcolor{teal}{\textbf{FAIR}} & \texttt{1.3.6} Amination of Heteroaryl Halides & & \texttt{1.3.8} Amination of Aryl Halides$^\dagger$ \\
  42 & 2 & \textcolor{teal}{\textbf{FAIR}} & \texttt{1.3.8} Amination of Aryl Halides & & \texttt{1.3.6} Amination of Heteroaryl Halides$^\dagger$ \\
  43 & 2 & \textcolor{teal}{\textbf{FAIR}} & \texttt{1.3.8} Amination of Aryl Halides & & \texttt{1.3.5} SNAr$^\dagger$ \\
  44 & 2 & \textcolor{gray}{\textbf{UNCLEAR}} & \texttt{1.3.Other}  & & \texttt{1.3.1} Buchwald-Hartwig$^\dagger$ \\
  45 & 2 & \textcolor{teal}{\textbf{FAIR}} & \texttt{1.6.1} N-Alkylation of $\pi$-Excessive Heterocycles & & \texttt{1.6.33} N-Alkylation with Benzyl Halides$^\dagger$ \\
  46 & 2 & \textcolor{teal}{\textbf{FAIR}} & \texttt{1.6.28} N-Alkylation with Secondary Alkyl Halides & & \texttt{1.6.1} N-Alkylation of $\pi$-Excessive Heterocycles (Pyrrole-type)$^\dagger$ \\
  47 & 2 & \textcolor{teal}{\textbf{FAIR}} & \texttt{1.6.29} N-Alkylation with Allyl Halides & & \texttt{1.6.1} N-Alkylation of $\pi$-Excessive Heterocycles (Pyrrole-type)$^\dagger$ \\
  48 & 2 & \textcolor{teal}{\textbf{FAIR}} & \texttt{1.7.5} O-Alkylation of COOH to form Esters & & \texttt{2.6.4} Esterification via Alkylation of Carboxylates \\
  49 & 2 & \textcolor{teal}{\textbf{FAIR}} & \texttt{1.8.2} Thia-Michael Addition & & \texttt{1.8.1} Thioether Formation$^\dagger$ \\
  50 & 2 & \textcolor{teal}{\textbf{FAIR}} & \texttt{2.1.1} Amidation using Acyl Halides & & \texttt{2.4.1} Carbamate Formation \\
  51 & 2 & \textcolor{red}{\textbf{SUSPECT}} & \texttt{2.1.2} Amidation using Carboxylic Acids & & \texttt{5.1.2} Cleavage of Amine PGs \\
  52 & 2 & \textcolor{teal}{\textbf{FAIR}} & \texttt{2.2.3} Sulfamide from Sulfamoyl Halides & & \texttt{2.2.1} Sulfonamide from Sulfonyl Halides$^\dagger$ \\
  53 & 2 & \textcolor{red}{\textbf{SUSPECT}} & \texttt{2.6.1} Esterification & & \texttt{5.1.4} Cleavage of COOH PGs \\
  54 & 2 & \textcolor{teal}{\textbf{FAIR}} & \texttt{2.7.6} O-sulfonylation of heteroaromatic lactams & & \texttt{2.7.3} Sulfonate ester from phenols$^\dagger$ \\
  55 & 2 & \textcolor{teal}{\textbf{FAIR}} & \texttt{3.11.113} Cyanation of heteroaryl methyl halides & & \texttt{3.11.22} Cyanation of benzyl halides$^\dagger$ \\
  56 & 2 & \textcolor{teal}{\textbf{FAIR}} & \texttt{3.11.16} Formylation of arenes & & \texttt{3.9.1} Organolithium Reactions \\
  57 & 2 & \textcolor{teal}{\textbf{FAIR}} & \texttt{3.5.8} Pd-Catalyzed Borylation & & \texttt{8.7.7} Synthesis of Organoboron Compounds \\
  58 & 2 & \textcolor{teal}{\textbf{FAIR}} & \texttt{4.1.4} Six-membered N-Heterocycle Synthesis & & \texttt{4.1.3} Five-membered N-Heterocycle Synthesis$^\dagger$ \\
  59 & 2 & \textcolor{teal}{\textbf{FAIR}} & \texttt{5.1.13} Decarboxylation and Decarbalkoxylation & & \texttt{8.7.8} Decarboxylation \\
  60 & 2 & \textcolor{teal}{\textbf{FAIR}} & \texttt{5.1.18} Simultaneous Deprotections & & \texttt{5.1.2} Cleavage of Amine PGs$^\dagger$ \\
  61 & 2 & \textcolor{teal}{\textbf{FAIR}} & \texttt{6.1.11} Reduction of nitrobenzenes to anilines & & \texttt{6.1.14} Reduction of nitroquinolines$^\dagger$ \\
  62 & 2 & \textcolor{teal}{\textbf{FAIR}} & \texttt{6.1.11} Reduction of nitrobenzenes to anilines & & \texttt{6.1.1} Catalytic Hydrogenation$^\dagger$ \\
  63 & 2 & \textcolor{teal}{\textbf{FAIR}} & \texttt{6.1.14} Reduction of nitroquinolines and fused nitropyridines to aminoquinolines and fused aminopyridines & & \texttt{6.1.11} Reduction of nitrobenzenes to anilines$^\dagger$ \\
  64 & 2 & \textcolor{teal}{\textbf{FAIR}} & \texttt{6.4.1} Reduction by Complex Metal Hydrides & & \texttt{5.1.1} Cleavage of Alcohol PGs \\
  65 & 2 & \textcolor{teal}{\textbf{FAIR}} & \texttt{6.6} Alkene to alkane & & \texttt{6.6.4} Reduction of styrenes \\
  66 & 2 & \textcolor{red}{\textbf{SUSPECT}} & \texttt{6.9.2} Reductive Cleavage & & \texttt{8.7.4} Amine Derivatizations \\
  67 & 2 & \textcolor{teal}{\textbf{FAIR}} & \texttt{8.6.5} Enaminone Formation & & \texttt{1.5.11} Condensation with activated vinyl ethers \\
  68 & 2 & \textcolor{teal}{\textbf{FAIR}} & \texttt{8.7.2} COOH Derivative Interconversions & & \texttt{8.4.2} Nitrile to Ester \\
  69 & 2 & \textcolor{red}{\textbf{SUSPECT}} & \texttt{8.7.4} Amine Derivatizations & & \texttt{1.3.6} Amination of Heteroaryl Halides \\
  70 & 2 & \textcolor{red}{\textbf{SUSPECT}} & \texttt{8.7.6} Miscellaneous FGIs & & \texttt{1.1.5} Alkylation with Alcohols \\
  71 & 2 & \textcolor{teal}{\textbf{FAIR}} & \texttt{9.1.5} Halogenation of Arenes & & \texttt{9.1.4} Halogenation of Heteroarenes$^\dagger$ \\
  72 & 2 & \textcolor{gray}{\textbf{UNCLEAR}} & \texttt{9.3.4} Chlorosulfonation of Arenes & & \texttt{9.7.5} ??? \\
  73 & 1 & \textcolor{red}{\textbf{SUSPECT}} & \texttt{1.1.1} SN2 & & \texttt{3.9.1} Organolithium Reactions \\
  74 & 1 & \textcolor{teal}{\textbf{FAIR}} & \texttt{1.1.1} SN2 & & \texttt{1.6.5} N-Alkylation of Pyridones \\
  75 & 1 & \textcolor{teal}{\textbf{FAIR}} & \texttt{1.1.26} Alkylation with Alkyl Sulfones & & \texttt{1.1.1} SN2$^\dagger$ \\
  76 & 1 & \textcolor{teal}{\textbf{FAIR}} & \texttt{1.1.27} N-cyanation with Cyanogen Halides & & \texttt{8.7.4} Amine Derivatizations \\
  77 & 1 & \textcolor{red}{\textbf{SUSPECT}} & \texttt{1.1.4} Aza-Michael Addition & & \texttt{6.6.41} Reduction of $\alpha,\beta$-unsaturated sulfonamides \\
  78 & 1 & \textcolor{red}{\textbf{SUSPECT}} & \texttt{1.2.1} Reductive Amination & & \texttt{1.3.5} SNAr \\
  79 & 1 & \textcolor{red}{\textbf{SUSPECT}} & \texttt{1.2.1} Reductive Amination & & \texttt{1.1.1} SN2 \\
  80 & 1 & \textcolor{teal}{\textbf{FAIR}} & \texttt{1.2.10} N-arylation with heteroaryl halides & & \texttt{1.3.1} Buchwald-Hartwig \\
  81 & 1 & \textcolor{teal}{\textbf{FAIR}} & \texttt{1.2.10} N-arylation with heteroaryl halides & & \texttt{1.3.6} Amination of Heteroaryl Halides \\
  82 & 1 & \textcolor{teal}{\textbf{FAIR}} & \texttt{1.2.11} N-arylation with heteroaryl sulfones & & \texttt{1.3.5} SNAr \\
  83 & 1 & \textcolor{teal}{\textbf{FAIR}} & \texttt{1.2.8} N-alkylation with benzyl halides & & \texttt{1.1.1} SN2 \\
  84 & 1 & \textcolor{red}{\textbf{SUSPECT}} & \texttt{1.3.1} Buchwald-Hartwig & & \texttt{1.1.1} SN2 \\
  85 & 1 & \textcolor{red}{\textbf{SUSPECT}} & \texttt{1.3.1} Buchwald-Hartwig & & \texttt{2.1.1} Amidation using Acyl Halides \\
  86 & 1 & \textcolor{teal}{\textbf{FAIR}} & \texttt{1.3.10} N-arylation of Cyclic Ureas & & \texttt{1.3.6} Amination of Heteroaryl Halides$^\dagger$ \\
  87 & 1 & \textcolor{teal}{\textbf{FAIR}} & \texttt{1.3.11} N-arylation with Arylboronic Acids & & \texttt{1.3.3} Chan-Lam Coupling$^\dagger$ \\
  88 & 1 & \textcolor{teal}{\textbf{FAIR}} & \texttt{1.3.15} Amination of Phenols & & \texttt{1.3.1} Buchwald-Hartwig$^\dagger$ \\
  89 & 1 & \textcolor{teal}{\textbf{FAIR}} & \texttt{1.3.19} N-arylation of Lactams & & \texttt{1.3.1} Buchwald-Hartwig$^\dagger$ \\
  90 & 1 & \textcolor{red}{\textbf{SUSPECT}} & \texttt{1.3.2} Ullmann & & \texttt{1.1.1} SN2 \\
  91 & 1 & \textcolor{teal}{\textbf{FAIR}} & \texttt{1.3.21} N-arylation of Sulfonamides & & \texttt{1.3.1} Buchwald-Hartwig$^\dagger$ \\
  92 & 1 & \textcolor{teal}{\textbf{FAIR}} & \texttt{1.3.23} Amination of Heteroaryl N-oxides & & \texttt{4.1.3} Five-membered N-Heterocycle Synthesis \\
  93 & 1 & \textcolor{teal}{\textbf{FAIR}} & \texttt{1.3.4} Goldberg & & \texttt{1.3.8} Amination of Aryl Halides$^\dagger$ \\
  94 & 1 & \textcolor{teal}{\textbf{FAIR}} & \texttt{1.3.4} Goldberg & & \texttt{1.3.19} N-arylation of Lactams$^\dagger$ \\
  95 & 1 & \textcolor{teal}{\textbf{FAIR}} & \texttt{1.3.40} Amination of 2-Halo-1,3-diazines & & \texttt{1.3.6} Amination of Heteroaryl Halides$^\dagger$ \\
  96 & 1 & \textcolor{teal}{\textbf{FAIR}} & \texttt{1.3.5} SNAr & & \texttt{1.3.19} N-arylation of Lactams$^\dagger$ \\
  97 & 1 & \textcolor{red}{\textbf{SUSPECT}} & \texttt{1.3.5} SNAr & & \texttt{5.1.2} Cleavage of Amine PGs \\
  98 & 1 & \textcolor{teal}{\textbf{FAIR}} & \texttt{1.3.6} Amination of Heteroaryl Halides & & \texttt{1.2.10} N-arylation with heteroaryl halides \\
  99 & 1 & \textcolor{red}{\textbf{SUSPECT}} & \texttt{1.3.6} Amination of Heteroaryl Halides & & \texttt{1.1.1} SN2 \\
  100 & 1 & \textcolor{gray}{\textbf{UNCLEAR}} & \texttt{1.3.8} Amination of Aryl Halides & & \texttt{1.3.Other} ??? \\
  101 & 1 & \textcolor{teal}{\textbf{FAIR}} & \texttt{1.3.8} Amination of Aryl Halides & & \texttt{1.3.2} Ullmann$^\dagger$ \\
  102 & 1 & \textcolor{teal}{\textbf{FAIR}} & \texttt{1.3.9} Amination of Heteroaryl-ones & & \texttt{1.3.6} Amination of Heteroaryl Halides$^\dagger$ \\
  103 & 1 & \textcolor{teal}{\textbf{FAIR}} & \texttt{1.4.1} N-Alkylation of Amides & & \texttt{1.1.1} SN2 \\
  104 & 1 & \textcolor{teal}{\textbf{FAIR}} & \texttt{1.4.16} N-Alkylation of Imidazolidinones & & \texttt{1.4.1} N-Alkylation of Amides$^\dagger$ \\
  105 & 1 & \textcolor{teal}{\textbf{FAIR}} & \texttt{1.4.21} N-Alkylation of Hydantoins & & \texttt{1.1.1} SN2 \\
  106 & 1 & \textcolor{teal}{\textbf{FAIR}} & \texttt{1.4.3} N-Alkylation of Imides & & \texttt{1.1.1} SN2 \\
  107 & 1 & \textcolor{teal}{\textbf{FAIR}} & \texttt{1.4.3} N-Alkylation of Imides & & \texttt{1.1.3} Alkylation with Epoxides \\
  108 & 1 & \textcolor{gray}{\textbf{UNCLEAR}} & \texttt{1.4.31} N-Alkylation of N-Benzylic Amides & & \texttt{1.4.Other} ??? \\
  109 & 1 & \textcolor{teal}{\textbf{FAIR}} & \texttt{1.4.4} N-Alkylation of Sulfonamides & & \texttt{1.1.5} Alkylation with Alcohols \\
  110 & 1 & \textcolor{teal}{\textbf{FAIR}} & \texttt{1.4.7} N-Alkylation of Cyclic Ureas & & \texttt{1.6.1} N-Alkylation of $\pi$-Excessive Heterocycles \\
  111 & 1 & \textcolor{red}{\textbf{SUSPECT}} & \texttt{1.4.7} N-Alkylation of Cyclic Ureas & & \texttt{1.3.6} Amination of Heteroaryl Halides \\
  112 & 1 & \textcolor{red}{\textbf{SUSPECT}} & \texttt{1.5.7} N-Alkylation with Alcohols & & \texttt{1.3.5} SNAr \\
  113 & 1 & \textcolor{teal}{\textbf{FAIR}} & \texttt{1.6.1} N-Alkylation of $\pi$-Excessive Heterocycles (Pyrrole-type) & & \texttt{1.6.5} N-Alkylation of Pyridones$^\dagger$ \\
  114 & 1 & \textcolor{teal}{\textbf{FAIR}} & \texttt{1.6.1} N-Alkylation of $\pi$-Excessive Heterocycles (Pyrrole-type) & & \texttt{1.6.13} N-Alkylation with Epoxides$^\dagger$ \\
  115 & 1 & \textcolor{teal}{\textbf{FAIR}} & \texttt{1.6.1} N-Alkylation of $\pi$-Excessive Heterocycles (Pyrrole-type) & & \texttt{1.1.3} Alkylation with Epoxides \\
  116 & 1 & \textcolor{red}{\textbf{SUSPECT}} & \texttt{1.6.1} N-Alkylation of $\pi$-Excessive Heterocycles & & \texttt{1.3.6} Amination of Heteroaryl Halides \\
  117 & 1 & \textcolor{teal}{\textbf{FAIR}} & \texttt{1.6.1} N-Alkylation of $\pi$-Excessive Heterocycles (Pyrrole-type) & & \texttt{1.1.1} SN2 \\
  118 & 1 & \textcolor{teal}{\textbf{FAIR}} & \texttt{1.6.11} N-Alkylation with Sulfonates & & \texttt{1.6.1} N-Alkylation of $\pi$-Excessive Heterocycles (Pyrrole-type)$^\dagger$ \\
  119 & 1 & \textcolor{teal}{\textbf{FAIR}} & \texttt{1.6.15} N-Alkylation with Halomethyl Heterocycles & & \texttt{1.6.1} N-Alkylation of $\pi$-Excessive Heterocycles (Pyrrole-type)$^\dagger$ \\
  120 & 1 & \textcolor{teal}{\textbf{FAIR}} & \texttt{1.6.16} N-Heteroarylation of N-Heterocycles & & \texttt{1.3.1} Buchwald-Hartwig \\
  121 & 1 & \textcolor{teal}{\textbf{FAIR}} & \texttt{1.6.19} N-Alkylation of Exocyclic Heteroarylamines & & \texttt{1.1.1} SN2 \\
  122 & 1 & \textcolor{teal}{\textbf{FAIR}} & \texttt{1.6.29} N-Alkylation with Allyl Halides & & \texttt{1.6.8} N-Alkylation with $\alpha$-Halo Carbonyls$^\dagger$ \\
  123 & 1 & \textcolor{teal}{\textbf{FAIR}} & \texttt{1.6.39} N-Alkylation with Methyl Halides & & \texttt{1.6.1} N-Alkylation of $\pi$-Excessive Heterocycles (Pyrrole-type)$^\dagger$ \\
  124 & 1 & \textcolor{red}{\textbf{SUSPECT}} & \texttt{1.6.39} N-Alkylation with Methyl Halides & & \texttt{1.3.6} Amination of Heteroaryl Halides \\
  125 & 1 & \textcolor{teal}{\textbf{FAIR}} & \texttt{1.6.5} N-Alkylation of Pyridones and Uracils & & \texttt{1.6.29} N-Alkylation with Allyl Halides$^\dagger$ \\
  126 & 1 & \textcolor{teal}{\textbf{FAIR}} & \texttt{1.6.5} N-Alkylation of Pyridones and Uracils & & \texttt{1.6.50} N-Alkylation with Primary Alkyl Halides$^\dagger$ \\
  127 & 1 & \textcolor{teal}{\textbf{FAIR}} & \texttt{1.6.6} N-Alkylation with Enol Ethers & & \texttt{1.5.8} Condensation with alkoxymethylene malonates \\
  128 & 1 & \textcolor{teal}{\textbf{FAIR}} & \texttt{1.6.7} N-Alkylation with $\alpha$-Haloethers & & \texttt{1.6.1} N-Alkylation of $\pi$-Excessive Heterocycles (Pyrrole-type)$^\dagger$ \\
  129 & 1 & \textcolor{teal}{\textbf{FAIR}} & \texttt{1.6.8} N-Alkylation with $\alpha$-Halo Carbonyls & & \texttt{1.6.1} N-Alkylation of $\pi$-Excessive Heterocycles (Pyrrole-type)$^\dagger$ \\
  130 & 1 & \textcolor{red}{\textbf{SUSPECT}} & \texttt{1.7.1} Williamson Ether Synthesis & & \texttt{8.7.2} COOH Derivative Interconversions \\
  131 & 1 & \textcolor{red}{\textbf{SUSPECT}} & \texttt{1.7.10} O-Alkylation with Diazo Compounds & & \texttt{5.1.1} Cleavage of Alcohol PGs \\
  132 & 1 & \textcolor{teal}{\textbf{FAIR}} & \texttt{1.7.12} O-Alkylation with Oxonium Salts & & \texttt{1.7.1} Williamson Ether Synthesis (Alkoxide + Organohalide/Sulfonate)$^\dagger$ \\
  133 & 1 & \textcolor{red}{\textbf{SUSPECT}} & \texttt{1.7.14} O-Glycosylation & & \texttt{2.6.1} Esterification \\
  134 & 1 & \textcolor{teal}{\textbf{FAIR}} & \texttt{1.7.18} O-Alkylation via Conjugate Addition & & \texttt{1.7.1} Williamson Ether Synthesis (Alkoxide + Organohalide/Sulfonate)$^\dagger$ \\
  135 & 1 & \textcolor{teal}{\textbf{FAIR}} & \texttt{1.7.23} O-Alkylation of Stabilized Enols & & \texttt{1.7.3} O-Alkylation using Alcohols as Electrophiles$^\dagger$ \\
  136 & 1 & \textcolor{red}{\textbf{SUSPECT}} & \texttt{1.7.26} Transetherification & & \texttt{6.8.1} Complete Hydrogenation of Aromatic Systems \\
  137 & 1 & \textcolor{teal}{\textbf{FAIR}} & \texttt{1.7.28} Hydrolysis of Primary Aromatic Amines & & \texttt{8.7.4} Amine Derivatizations \\
  138 & 1 & \textcolor{teal}{\textbf{FAIR}} & \texttt{1.7.30} O-Methylation with DMF & & \texttt{2.6.18} Esterification with DMF \\
  139 & 1 & \textcolor{teal}{\textbf{FAIR}} & \texttt{1.7.4} Ring-opening of Epoxides with O-Nucleophiles & & \texttt{1.7.3} O-Alkylation using Alcohols as Electrophiles$^\dagger$ \\
  140 & 1 & \textcolor{teal}{\textbf{FAIR}} & \texttt{1.7.4} Ring-opening of Epoxides & & \texttt{2.6.1} Esterification \\
  141 & 1 & \textcolor{teal}{\textbf{FAIR}} & \texttt{1.7.6} O-Silylation & & \texttt{1.7.2} O-Arylation$^\dagger$ \\
  142 & 1 & \textcolor{red}{\textbf{SUSPECT}} & \texttt{1.7.7} O-Alkylation for PG Formation & & \texttt{8.7.6} Miscellaneous FGIs \\
  143 & 1 & \textcolor{red}{\textbf{SUSPECT}} & \texttt{1.7.7} O-Alkylation for PG Formation & & \texttt{5.1.3} Cleavage of Carbonyl PGs \\
  144 & 1 & \textcolor{teal}{\textbf{FAIR}} & \texttt{1.7.8} Cleavage of Ether PGs & & \texttt{5.1.1} Cleavage of Alcohol PGs \\
  145 & 1 & \textcolor{red}{\textbf{SUSPECT}} & \texttt{2.1.1} Amidation using Acyl Halides & & \texttt{1.3.6} Amination of Heteroaryl Halides \\
  146 & 1 & \textcolor{teal}{\textbf{FAIR}} & \texttt{2.1.2} Amidation & & \texttt{4.1.3} Five-membered N-Heterocycle Synthesis \\
  147 & 1 & \textcolor{teal}{\textbf{FAIR}} & \texttt{2.1.2} Amidation & & \texttt{8.7.2} COOH Derivative Interconversions \\
  148 & 1 & \textcolor{teal}{\textbf{FAIR}} & \texttt{2.1.3} Amidation using Anhydrides & & \texttt{2.6.3} Acylation of Alcohols with Anhydrides \\
  149 & 1 & \textcolor{teal}{\textbf{FAIR}} & \texttt{2.1.3} Amidation using Anhydrides & & \texttt{2.4.1} Carbamate Formation \\
  150 & 1 & \textcolor{teal}{\textbf{FAIR}} & \texttt{2.1.3} Amidation using Anhydrides & & \texttt{4.1.4} Six-membered N-Heterocycle Synthesis \\
  151 & 1 & \textcolor{teal}{\textbf{FAIR}} & \texttt{2.1.3} Amidation using Anhydrides & & \texttt{2.1.2} Amidation using Carboxylic Acids$^\dagger$ \\
  152 & 1 & \textcolor{teal}{\textbf{FAIR}} & \texttt{2.1.4} Amidation using Esters & & \texttt{4.1.3} Five-membered N-Heterocycle Synthesis \\
  153 & 1 & \textcolor{red}{\textbf{SUSPECT}} & \texttt{2.1.5} Other N-Acylation & & \texttt{1.3.5} SNAr \\
  154 & 1 & \textcolor{teal}{\textbf{FAIR}} & \texttt{2.1.5} Other N-Acylation & & \texttt{8.7.2} COOH Derivative Interconversions \\
  155 & 1 & \textcolor{teal}{\textbf{FAIR}} & \texttt{2.1.5} Other N-Acylation & & \texttt{1.3.13} Amination of Squarate Esters \\
  156 & 1 & \textcolor{teal}{\textbf{FAIR}} & \texttt{2.1.6} Transamidation & & \texttt{2.1.2} Amidation using Carboxylic Acids$^\dagger$ \\
  157 & 1 & \textcolor{teal}{\textbf{FAIR}} & \texttt{2.2.5} Sulfonamide from Sulfonic Acids & & \texttt{2.1.2} Amidation \\
  158 & 1 & \textcolor{teal}{\textbf{FAIR}} & \texttt{2.3.20} Urea from Ureas & & \texttt{2.1.2} Amidation \\
  159 & 1 & \textcolor{teal}{\textbf{FAIR}} & \texttt{2.3.4} Urea from Carbamates & & \texttt{5.1.2} Cleavage of Amine PGs \\
  160 & 1 & \textcolor{teal}{\textbf{FAIR}} & \texttt{2.3.8} Urea from Carbamic Acids & & \texttt{2.1.2} Amidation \\
  161 & 1 & \textcolor{red}{\textbf{SUSPECT}} & \texttt{2.4.1} Carbamate Formation & & \texttt{5.1.22} Cleavage of Oxazolidinones \\
  162 & 1 & \textcolor{teal}{\textbf{FAIR}} & \texttt{2.4.3} Carbamate Cleavage & & \texttt{5.1.2} Cleavage of Amine PGs \\
  163 & 1 & \textcolor{red}{\textbf{SUSPECT}} & \texttt{2.5.13} Isothiourea Formation & & \texttt{1.3.5} SNAr \\
  164 & 1 & \textcolor{red}{\textbf{SUSPECT}} & \texttt{2.5.2} Amidine from Imidoyl Chlorides & & \texttt{1.3.6} Amination of Heteroaryl Halides \\
  165 & 1 & \textcolor{teal}{\textbf{FAIR}} & \texttt{2.5.3} Amidine from Amide/Thioamide & & \texttt{2.1.1} Amidation using Acyl Halides \\
  166 & 1 & \textcolor{teal}{\textbf{FAIR}} & \texttt{2.5.6} Amidine from Imidates & & \texttt{2.1.4} Amidation using Esters \\
  167 & 1 & \textcolor{teal}{\textbf{FAIR}} & \texttt{2.6.1} Esterification & & \texttt{2.1.2} Amidation \\
  168 & 1 & \textcolor{teal}{\textbf{FAIR}} & \texttt{2.6.2} Acylation of Alcohols with Acyl Halides & & \texttt{2.1.1} Amidation using Acyl Halides \\
  169 & 1 & \textcolor{teal}{\textbf{FAIR}} & \texttt{2.6.4} Esterification via Alkylation & & \texttt{5.1.4} Cleavage of COOH PGs \\
  170 & 1 & \textcolor{teal}{\textbf{FAIR}} & \texttt{2.6.4} Esterification via Alkylation of Carboxylates & & \texttt{2.1.2} Amidation \\
  171 & 1 & \textcolor{teal}{\textbf{FAIR}} & \texttt{2.8.13} Isothiocyanate Formation & & \texttt{8.7.4} Amine Derivatizations \\
  172 & 1 & \textcolor{red}{\textbf{SUSPECT}} & \texttt{2.8.15} Anhydride Formation & & \texttt{3.6.5} Condensation of COOH with Malonic Acid \\
  173 & 1 & \textcolor{teal}{\textbf{FAIR}} & \texttt{2.8.15} Anhydride Formation & & \texttt{2.4.2} Carbonate Formation \\
  174 & 1 & \textcolor{teal}{\textbf{FAIR}} & \texttt{2.8.2} S-Acylation & & \texttt{2.1.2} Amidation \\
  175 & 1 & \textcolor{teal}{\textbf{FAIR}} & \texttt{2.8.8} Isocyanate Formation & & \texttt{8.7.4} Amine Derivatizations \\
  176 & 1 & \textcolor{red}{\textbf{SUSPECT}} & \texttt{3.1.1} C(sp$^2$)-C(sp$^2$) Coupling & & \texttt{3.5.11} Reductive Coupling \\
  177 & 1 & \textcolor{red}{\textbf{SUSPECT}} & \texttt{3.1.1} C(sp$^2$)-C(sp$^2$) Coupling & & \texttt{1.3.1} Buchwald-Hartwig \\
  178 & 1 & \textcolor{teal}{\textbf{FAIR}} & \texttt{3.1.3} C(sp$^3$)-C(sp$^2$) Coupling & & \texttt{3.1.1} C(sp²)-C(sp²) Coupling (e.g., Biaryl or Styrene Synthesis)$^\dagger$ \\
  179 & 1 & \textcolor{red}{\textbf{SUSPECT}} & \texttt{3.10.1} Friedel-Crafts Acylation & & \texttt{8.4.1} Nitrile to Carboxylic Acid \\
  180 & 1 & \textcolor{red}{\textbf{SUSPECT}} & \texttt{3.11.1} Aldol Reactions & & \texttt{1.7.1} Williamson Ether \\
  181 & 1 & \textcolor{red}{\textbf{SUSPECT}} & \texttt{3.11.14} Condensation of aliphatic aldehydes with active methylene & & \texttt{2.1.2} Amidation \\
  182 & 1 & \textcolor{teal}{\textbf{FAIR}} & \texttt{3.11.16} Formylation of arenes & & \texttt{3.11.9} Formylation of phenols$^\dagger$ \\
  183 & 1 & \textcolor{red}{\textbf{SUSPECT}} & \texttt{3.11.17} Alkylation of arenes & & \texttt{6.9.1} Deoxygenation \\
  184 & 1 & \textcolor{teal}{\textbf{FAIR}} & \texttt{3.11.22} Cyanation of benzyl halides & & \texttt{3.11.11} Cyanation of alkyl halides$^\dagger$ \\
  185 & 1 & \textcolor{teal}{\textbf{FAIR}} & \texttt{3.11.23} Formylation of aryl halides & & \texttt{3.9.1} Organolithium Reactions \\
  186 & 1 & \textcolor{teal}{\textbf{FAIR}} & \texttt{3.11.25} Alkylation of active methylene & & \texttt{1.1.1} SN2 \\
  187 & 1 & \textcolor{teal}{\textbf{FAIR}} & \texttt{3.11.26} Alkylation of carboxylic acids & & \texttt{1.7.5} O-Alkylation of COOH to form Esters \\
  188 & 1 & \textcolor{teal}{\textbf{FAIR}} & \texttt{3.11.3} Mannich Reaction & & \texttt{1.2.6} Mannich Reaction \\
  189 & 1 & \textcolor{red}{\textbf{SUSPECT}} & \texttt{3.11.35} Hydrosilylation & & \texttt{4.1.3} Five-membered N-Heterocycle Synthesis \\
  190 & 1 & \textcolor{teal}{\textbf{FAIR}} & \texttt{3.11.38} Acylation of active methylene & & \texttt{3.9.1} Organolithium Reactions \\
  191 & 1 & \textcolor{teal}{\textbf{FAIR}} & \texttt{3.11.5} Aromatic Aldehyde Condensations & & \texttt{6.2.4} Condensation of oxindoles \\
  192 & 1 & \textcolor{teal}{\textbf{FAIR}} & \texttt{3.11.53} Alkylation of esters & & \texttt{3.11.29} Alkylation of amides and imides$^\dagger$ \\
  193 & 1 & \textcolor{red}{\textbf{SUSPECT}} & \texttt{3.11.6} Olefin Metathesis & & \texttt{6.6.1} Catalytic Hydrogenation \\
  194 & 1 & \textcolor{red}{\textbf{SUSPECT}} & \texttt{3.11.6} Olefin Metathesis & & \texttt{5.1.1} Cleavage of Alcohol PGs \\
  195 & 1 & \textcolor{red}{\textbf{SUSPECT}} & \texttt{3.11.9} Formylation of phenols & & \texttt{1.2.1} Reductive Amination \\
  196 & 1 & \textcolor{teal}{\textbf{FAIR}} & \texttt{3.2.1} Standard Heck reaction & & \texttt{3.2} Heck Reaction \\
  197 & 1 & \textcolor{teal}{\textbf{FAIR}} & \texttt{3.3.1} Coupling of Aryl/Vinyl Halides & & \texttt{3.3.5} Coupling of Heteroaryl Halides$^\dagger$ \\
  198 & 1 & \textcolor{teal}{\textbf{FAIR}} & \texttt{3.5.1} Negishi Coupling & & \texttt{3.9.1} Organolithium Reactions \\
  199 & 1 & \textcolor{red}{\textbf{SUSPECT}} & \texttt{3.5.2} Kumada Coupling & & \texttt{3.7.1} 1,2-Addition to Carbonyls \\
  200 & 1 & \textcolor{red}{\textbf{SUSPECT}} & \texttt{3.6.2} Mixed Claisen Condensation & & \texttt{3.11.29} Alkylation of amides and imides \\
  201 & 1 & \textcolor{red}{\textbf{SUSPECT}} & \texttt{3.6.2} Mixed Claisen Condensation & & \texttt{3.9.1} Organolithium Reactions \\
  202 & 1 & \textcolor{teal}{\textbf{FAIR}} & \texttt{3.6.3} Dieckmann Condensation & & \texttt{2.1.4} Amidation using Esters \\
  203 & 1 & \textcolor{teal}{\textbf{FAIR}} & \texttt{3.6.7} Condensation of malonate esters with acyl chlorides & & \texttt{3.11.38} Acylation of active methylene \\
  204 & 1 & \textcolor{red}{\textbf{SUSPECT}} & \texttt{3.7.4} Nucleophilic Aromatic Substitution & & \texttt{3.5.2} Kumada Coupling \\
  205 & 1 & \textcolor{gray}{\textbf{UNCLEAR}} & \texttt{3.9.1} Organolithium & & \texttt{6.Other.1} ??? \\
  206 & 1 & \textcolor{red}{\textbf{SUSPECT}} & \texttt{3.9.1} Organolithium & & \texttt{4.1.3} Five-membered N-Heterocycle Synthesis \\
  207 & 1 & \textcolor{red}{\textbf{SUSPECT}} & \texttt{3.9.1} Organolithium & & \texttt{1.7.1} Williamson Ether Synthesis \\
  208 & 1 & \textcolor{red}{\textbf{SUSPECT}} & \texttt{3.9.1} Organolithium & & \texttt{6.9.1} Deoxygenation \\
  209 & 1 & \textcolor{teal}{\textbf{FAIR}} & \texttt{3.9.1} Organolithium & & \texttt{3.6.2} Mixed Claisen Condensation \\
  210 & 1 & \textcolor{teal}{\textbf{FAIR}} & \texttt{3.9.1} Organolithium & & \texttt{3.11.16} Formylation of arenes \\
  211 & 1 & \textcolor{teal}{\textbf{FAIR}} & \texttt{4.1.3} Five-membered N-Heterocycle Synthesis & & \texttt{2.1.3} Amidation using Anhydrides \\
  212 & 1 & \textcolor{teal}{\textbf{FAIR}} & \texttt{4.1.3} Five-membered N-Heterocycle Synthesis & & \texttt{2.3.2} Urea from Phosgene \\
  213 & 1 & \textcolor{red}{\textbf{SUSPECT}} & \texttt{4.1.4} Six-membered N-Heterocycle Synthesis & & \texttt{1.3.6} Amination of Heteroaryl Halides \\
  214 & 1 & \textcolor{teal}{\textbf{FAIR}} & \texttt{4.1.4} Six-membered N-Heterocycle Synthesis & & \texttt{2.1.3} Amidation using Anhydrides \\
  215 & 1 & \textcolor{teal}{\textbf{FAIR}} & \texttt{4.1.5} Seven-membered N-Heterocycle Synthesis & & \texttt{4.1.3} Five-membered$^\dagger$ \\
  216 & 1 & \textcolor{teal}{\textbf{FAIR}} & \texttt{4.2.1} Epoxide Synthesis & & \texttt{7.7.1} Epoxidation \\
  217 & 1 & \textcolor{teal}{\textbf{FAIR}} & \texttt{5.1.1} Cleavage of Alcohol PGs & & \texttt{5.1.9} Cleavage of Phenol PGs$^\dagger$ \\
  218 & 1 & \textcolor{teal}{\textbf{FAIR}} & \texttt{5.1.1} Cleavage of Alcohol PGs & & \texttt{6.4.1} Reduction by Complex Metal Hydrides \\
  219 & 1 & \textcolor{red}{\textbf{SUSPECT}} & \texttt{5.1.1} Cleavage of Alcohol PGs & & \texttt{1.7.1} Williamson Ether Synthesis \\
  220 & 1 & \textcolor{red}{\textbf{SUSPECT}} & \texttt{5.1.13} Decarboxylation & & \texttt{3.11.28} Alkylation of nitriles \\
  221 & 1 & \textcolor{teal}{\textbf{FAIR}} & \texttt{5.1.13} Decarboxylation & & \texttt{8.7.2} COOH Derivative Interconversions \\
  222 & 1 & \textcolor{red}{\textbf{SUSPECT}} & \texttt{5.1.13} Decarboxylation & & \texttt{2.1.4} Amidation using Esters \\
  223 & 1 & \textcolor{teal}{\textbf{FAIR}} & \texttt{5.1.15} Cleavage of Thiocarboxylic Acid PGs & & \texttt{8.7.2} COOH Derivative Interconversions \\
  224 & 1 & \textcolor{teal}{\textbf{FAIR}} & \texttt{5.1.18} Simultaneous Deprotections & & \texttt{6.1} NO$_2$ to amine \\
  225 & 1 & \textcolor{teal}{\textbf{FAIR}} & \texttt{5.1.2} Cleavage of Amine PGs & & \texttt{8.7.2} COOH Derivative Interconversions \\
  226 & 1 & \textcolor{red}{\textbf{SUSPECT}} & \texttt{5.1.2} Cleavage of Amine PGs & & \texttt{1.3.5} SNAr \\
  227 & 1 & \textcolor{red}{\textbf{SUSPECT}} & \texttt{5.1.2} Cleavage of Amine PGs & & \texttt{2.1.2} Amidation \\
  228 & 1 & \textcolor{red}{\textbf{SUSPECT}} & \texttt{5.1.2} Cleavage of Amine PGs & & \texttt{2.3.1} Urea from Isocyanates \\
  229 & 1 & \textcolor{red}{\textbf{SUSPECT}} & \texttt{5.1.2} Cleavage of Amine PGs & & \texttt{1.2.1} Reductive Amination \\
  230 & 1 & \textcolor{red}{\textbf{SUSPECT}} & \texttt{5.1.9} Cleavage of Phenol PGs & & \texttt{1.7.1} Williamson Ether \\
  231 & 1 & \textcolor{red}{\textbf{SUSPECT}} & \texttt{5.1.9} Cleavage of Phenol PGs & & \texttt{4.1.4} Six-membered N-Heterocycle Synthesis \\
  232 & 1 & \textcolor{teal}{\textbf{FAIR}} & \texttt{6.1.11} Reduction of nitrobenzenes to anilines & & \texttt{6.1.16} Reduction of nitrobenzylamines to aminobenzylamines$^\dagger$ \\
  233 & 1 & \textcolor{teal}{\textbf{FAIR}} & \texttt{6.1.11} Reduction of nitrobenzenes to anilines & & \texttt{6.1} NO$_2$ to amine \\
  234 & 1 & \textcolor{red}{\textbf{SUSPECT}} & \texttt{6.1.11} Reduction of nitrobenzenes & & \texttt{1.3.5} SNAr \\
  235 & 1 & \textcolor{teal}{\textbf{FAIR}} & \texttt{6.1.18} Reductive alkoxycarbonylation of nitro & & \texttt{2.4.1} Carbamate Formation \\
  236 & 1 & \textcolor{teal}{\textbf{FAIR}} & \texttt{6.1.24} Reduction of nitrobenzyl ethers & & \texttt{6.1.11} Reduction of nitrobenzenes to anilines$^\dagger$ \\
  237 & 1 & \textcolor{teal}{\textbf{FAIR}} & \texttt{6.1.5} Reduction of nitropyrazoles & & \texttt{6.1.11} Reduction of nitrobenzenes to anilines$^\dagger$ \\
  238 & 1 & \textcolor{teal}{\textbf{FAIR}} & \texttt{6.1.8} Reduction of nitrobenzoic acid esters & & \texttt{6.1.1} Catalytic Hydrogenation$^\dagger$ \\
  239 & 1 & \textcolor{teal}{\textbf{FAIR}} & \texttt{6.1.8} Reduction of nitrobenzoic acid esters to aminobenzoic acid esters & & \texttt{6.1.11} Reduction of nitrobenzenes to anilines$^\dagger$ \\
  240 & 1 & \textcolor{teal}{\textbf{FAIR}} & \texttt{6.1.8} Reduction of nitrobenzoic acid esters to aminobenzoic acid esters & & \texttt{6.1.16} Reduction of nitrobenzylamines to aminobenzylamines$^\dagger$ \\
  241 & 1 & \textcolor{teal}{\textbf{FAIR}} & \texttt{6.2.1} Complex Metal Hydride Reduction & & \texttt{6.2.7} Reduction of cyclic imides to cyclic amines$^\dagger$ \\
  242 & 1 & \textcolor{teal}{\textbf{FAIR}} & \texttt{6.2.1} Complex Metal Hydride Reduction & & \texttt{5.1.2} Cleavage of Amine PGs \\
  243 & 1 & \textcolor{teal}{\textbf{FAIR}} & \texttt{6.2.14} Reduction of N-Aryl Amides to N-Aryl Amines & & \texttt{6.2.1} Complex Metal Hydride Reduction$^\dagger$ \\
  244 & 1 & \textcolor{teal}{\textbf{FAIR}} & \texttt{6.2.7} Reduction of cyclic imides to cyclic amines & & \texttt{6.2.1} Complex Metal Hydride Reduction$^\dagger$ \\
  245 & 1 & \textcolor{red}{\textbf{SUSPECT}} & \texttt{6.3.1} Nitrile Reduction & & \texttt{5.1.2} Cleavage of Amine PGs \\
  246 & 1 & \textcolor{teal}{\textbf{FAIR}} & \texttt{6.4.1} Reduction by Complex Metal Hydrides & & \texttt{6.9.1} Deoxygenation \\
  247 & 1 & \textcolor{teal}{\textbf{FAIR}} & \texttt{6.5.1} Hydride Reduction & & \texttt{6.5.12} Reduction of $\alpha,\beta$-unsaturated ketones$^\dagger$ \\
  248 & 1 & \textcolor{teal}{\textbf{FAIR}} & \texttt{6.5.10} Reduction of amino ketones & & \texttt{6.5.1} Hydride Reduction$^\dagger$ \\
  249 & 1 & \textcolor{teal}{\textbf{FAIR}} & \texttt{6.5.12} Reduction of $\alpha,\beta$-unsaturated ketones & & \texttt{6.5.1} Hydride Reduction$^\dagger$ \\
  250 & 1 & \textcolor{red}{\textbf{SUSPECT}} & \texttt{6.5.5} Reduction of aryl ketones & & \texttt{7.1.2} Oxidation of Secondary Alcohols to Ketones \\
  251 & 1 & \textcolor{red}{\textbf{SUSPECT}} & \texttt{6.5.5} Reduction of aryl ketones & & \texttt{4.1.3} Five-membered N-Heterocycle Synthesis \\
  252 & 1 & \textcolor{teal}{\textbf{FAIR}} & \texttt{6.5.6} Reduction of $\beta$-keto esters & & \texttt{6.5.1} Hydride Reduction$^\dagger$ \\
  253 & 1 & \textcolor{teal}{\textbf{FAIR}} & \texttt{6.6} Alkene to alkane & & \texttt{6.6.1} Catalytic Hydrogenation \\
  254 & 1 & \textcolor{teal}{\textbf{FAIR}} & \texttt{6.6.1} Catalytic Hydrogenation & & \texttt{6.6.3} Reduction of Tetrahydropyridines$^\dagger$ \\
  255 & 1 & \textcolor{teal}{\textbf{FAIR}} & \texttt{6.6.1} Catalytic Hydrogenation & & \texttt{6.6.5} Reduction of cinnamates$^\dagger$ \\
  256 & 1 & \textcolor{red}{\textbf{SUSPECT}} & \texttt{6.6.1} Catalytic Hydrogenation & & \texttt{7.1.2} Oxidation of Secondary Alcohols to Ketones \\
  257 & 1 & \textcolor{teal}{\textbf{FAIR}} & \texttt{6.6.10} Reduction of benzylidene ketones & & \texttt{6.6.1} Catalytic Hydrogenation$^\dagger$ \\
  258 & 1 & \textcolor{teal}{\textbf{FAIR}} & \texttt{6.6.12} Reduction of N-heteroarylalkenes & & \texttt{6.6.39} Reduction of dihydropyrans$^\dagger$ \\
  259 & 1 & \textcolor{teal}{\textbf{FAIR}} & \texttt{6.6.14} Reduction of $\alpha,\beta$-unsat esters & & \texttt{6.6.5} Reduction of cinnamates$^\dagger$ \\
  260 & 1 & \textcolor{teal}{\textbf{FAIR}} & \texttt{6.6.16} Reduction of cinnamic acids & & \texttt{6.6.1} Catalytic Hydrogenation$^\dagger$ \\
  261 & 1 & \textcolor{teal}{\textbf{FAIR}} & \texttt{6.6.27} Reduction of internal alkenes & & \texttt{6.4.1} Reduction by Complex Metal Hydrides \\
  262 & 1 & \textcolor{red}{\textbf{SUSPECT}} & \texttt{6.6.30} Reduction of $\alpha,\beta$-unsat amides & & \texttt{6.8.1} Complete Hydrogenation of Aromatic Systems \\
  263 & 1 & \textcolor{teal}{\textbf{FAIR}} & \texttt{6.6.4} Reduction of styrenes & & \texttt{6.6} Alkene to alkane \\
  264 & 1 & \textcolor{red}{\textbf{SUSPECT}} & \texttt{6.6.4} Reduction of styrenes & & \texttt{6.9.1} Deoxygenation \\
  265 & 1 & \textcolor{teal}{\textbf{FAIR}} & \texttt{6.6.4} Reduction of styrenes to alkylarenes & & \texttt{6.5.1} Hydride Reduction \\
  266 & 1 & \textcolor{teal}{\textbf{FAIR}} & \texttt{6.6.7} Reduction of terminal alkenes & & \texttt{6.6.1} Catalytic Hydrogenation$^\dagger$ \\
  267 & 1 & \textcolor{teal}{\textbf{FAIR}} & \texttt{6.8.1} Complete Hydrogenation of Aromatic Systems & & \texttt{6.9.12} Reduction of Aromatic Rings \\
  268 & 1 & \textcolor{red}{\textbf{SUSPECT}} & \texttt{6.9.1} Deoxygenation & & \texttt{7.8.1} Oxidation of Carbonyl Compounds \\
  269 & 1 & \textcolor{red}{\textbf{SUSPECT}} & \texttt{6.9.1} Deoxygenation & & \texttt{1.3.8} Amination of Aryl Halides \\
  270 & 1 & \textcolor{red}{\textbf{SUSPECT}} & \texttt{6.9.1} Deoxygenation & & \texttt{1.2.1} Reductive Amination \\
  271 & 1 & \textcolor{red}{\textbf{SUSPECT}} & \texttt{6.9.1} Deoxygenation & & \texttt{2.1.2} Amidation \\
  272 & 1 & \textcolor{teal}{\textbf{FAIR}} & \texttt{6.9.1} Deoxygenation of Alcohols and Carbonyls & & \texttt{6.9.2} Reductive Cleavage$^\dagger$ \\
  273 & 1 & \textcolor{red}{\textbf{SUSPECT}} & \texttt{6.9.1} Deoxygenation & & \texttt{1.7.2} O-Arylation \\
  274 & 1 & \textcolor{teal}{\textbf{FAIR}} & \texttt{6.9.10} Reduction of Acyl Halides & & \texttt{6.9.8} Reduction of Esters$^\dagger$ \\
  275 & 1 & \textcolor{red}{\textbf{SUSPECT}} & \texttt{6.9.12} Reduction of Aromatic Rings & & \texttt{5.1.4} Cleavage of COOH PGs \\
  276 & 1 & \textcolor{teal}{\textbf{FAIR}} & \texttt{6.9.2} Reductive Cleavage & & \texttt{6.9.1} Deoxygenation of Alcohols and Carbonyls$^\dagger$ \\
  277 & 1 & \textcolor{red}{\textbf{SUSPECT}} & \texttt{6.9.2} Reductive Cleavage & & \texttt{7.8.1} Oxidation of Carbonyl Compounds \\
  278 & 1 & \textcolor{red}{\textbf{SUSPECT}} & \texttt{6.9.2} Reductive Cleavage & & \texttt{1.1.3} Alkylation with Epoxides \\
  279 & 1 & \textcolor{teal}{\textbf{FAIR}} & \texttt{6.9.2} Reductive Cleavage and Decarboxylation & & \texttt{5.1.1} Cleavage of Alcohol PGs \\
  280 & 1 & \textcolor{teal}{\textbf{FAIR}} & \texttt{6.9.2} Reductive Cleavage and Decarboxylation & & \texttt{8.7.2} COOH Derivative Interconversions \\
  281 & 1 & \textcolor{red}{\textbf{SUSPECT}} & \texttt{6.9.2} Reductive Cleavage & & \texttt{1.7.1} Williamson Ether \\
  282 & 1 & \textcolor{teal}{\textbf{FAIR}} & \texttt{6.9.3} Reduction of N-Containing FGs & & \texttt{5.1.2} Cleavage of Amine PGs \\
  283 & 1 & \textcolor{teal}{\textbf{FAIR}} & \texttt{6.9.3} Reduction of Nitrogen-Containing Functional Groups & & \texttt{8.7.4} Amine Derivatizations \\
  284 & 1 & \textcolor{red}{\textbf{SUSPECT}} & \texttt{6.9.5} Reduction of Sulfur FGs & & \texttt{1.2.11} N-arylation with heteroaryl sulfones \\
  285 & 1 & \textcolor{red}{\textbf{SUSPECT}} & \texttt{6.9.5} Reduction of Sulfur Functional Groups & & \texttt{1.3.2} Ullmann Condensation \\
  286 & 1 & \textcolor{gray}{\textbf{UNCLEAR}} & \texttt{6.Other.1} ??? & & \texttt{6.9.11} Reduction of Aldehydes \\
  287 & 1 & \textcolor{red}{\textbf{SUSPECT}} & \texttt{7.1.2} Oxidation of Secondary Alcohols & & \texttt{6.9.1} Deoxygenation \\
  288 & 1 & \textcolor{red}{\textbf{SUSPECT}} & \texttt{7.1.2} Oxidation of Secondary Alcohols & & \texttt{1.7.3} O-Alkylation using Alcohols as Electrophiles \\
  289 & 1 & \textcolor{teal}{\textbf{FAIR}} & \texttt{7.2.2} Oxidation of Sulfoxides to Sulfones & & \texttt{7.2.1} Oxidation of Sulfides$^\dagger$ \\
  290 & 1 & \textcolor{teal}{\textbf{FAIR}} & \texttt{7.6.1} Oxidation of Benzylic C-H Bonds & & \texttt{3.11.16} Formylation of arenes \\
  291 & 1 & \textcolor{teal}{\textbf{FAIR}} & \texttt{7.6.3} Oxidation of Benzylic/Allylic Alcohols & & \texttt{7.8.7} Oxidation of Alcohols \\
  292 & 1 & \textcolor{teal}{\textbf{FAIR}} & \texttt{7.6.4} Halogenation of Benzylic C-H Bonds & & \texttt{9.1.6} Benzylic Halogenation \\
  293 & 1 & \textcolor{red}{\textbf{SUSPECT}} & \texttt{7.8.1} Oxidation of Carbonyl Compounds & & \texttt{6.9.6} Reduction of Carboxylic Acids \\
  294 & 1 & \textcolor{red}{\textbf{SUSPECT}} & \texttt{7.8.1} Oxidation of Carbonyl Compounds & & \texttt{2.6.1} Esterification \\
  295 & 1 & \textcolor{red}{\textbf{SUSPECT}} & \texttt{7.8.7} Oxidation of Alcohols & & \texttt{5.1.3} Cleavage of Carbonyl PGs \\
  296 & 1 & \textcolor{red}{\textbf{SUSPECT}} & \texttt{7.8.9} Oxidation of N-Heterocycles & & \texttt{6.8.1} Complete Hydrogenation of Aromatic Systems \\
  297 & 1 & \textcolor{red}{\textbf{SUSPECT}} & \texttt{7.8.9} Oxidation of N-Heterocycles & & \texttt{1.3.5} SNAr \\
  298 & 1 & \textcolor{teal}{\textbf{FAIR}} & \texttt{8.1.5} Heteroaromatic Alcohol to Heteroaryl Chloride & & \texttt{8.2.4} Deoxychlorination of Lactams \\
  299 & 1 & \textcolor{teal}{\textbf{FAIR}} & \texttt{8.4.1} Nitrile to Carboxylic Acid & & \texttt{8.7.2} COOH Derivative Interconversions \\
  300 & 1 & \textcolor{red}{\textbf{SUSPECT}} & \texttt{8.5.1} Dehydration of Alcohols to Alkenes & & \texttt{7.1.2} Oxidation of Secondary Alcohols \\
  301 & 1 & \textcolor{red}{\textbf{SUSPECT}} & \texttt{8.5.1} Dehydration & & \texttt{1.1.4} Aza-Michael Addition \\
  302 & 1 & \textcolor{teal}{\textbf{FAIR}} & \texttt{8.6.18} Imidate Formation & & \texttt{8.7.2} COOH Derivative Interconversions \\
  303 & 1 & \textcolor{teal}{\textbf{FAIR}} & \texttt{8.6.4} Semicarbazone Formation & & \texttt{8.6.8} Acyl Hydrazone Formation$^\dagger$ \\
  304 & 1 & \textcolor{teal}{\textbf{FAIR}} & \texttt{8.6.5} Enaminone Formation & & \texttt{1.5.8} Condensation with alkoxymethylene malonates \\
  305 & 1 & \textcolor{teal}{\textbf{FAIR}} & \texttt{8.7.10} Synthesis of Organotin Compounds & & \texttt{3.11.140} Stannylation of arenes \\
  306 & 1 & \textcolor{teal}{\textbf{FAIR}} & \texttt{8.7.2} COOH Derivative Interconversions & & \texttt{5.1.4} Cleavage of COOH PGs \\
  307 & 1 & \textcolor{teal}{\textbf{FAIR}} & \texttt{8.7.2} Carboxylic Acid Derivative Interconversions & & \texttt{8.6.9} Amidoxime Formation \\
  308 & 1 & \textcolor{teal}{\textbf{FAIR}} & \texttt{8.7.3} Rearrangements & & \texttt{2.1.2} Amidation \\
  309 & 1 & \textcolor{teal}{\textbf{FAIR}} & \texttt{8.7.4} Amine Derivatizations & & \texttt{2.8.13} Isothiocyanate Formation \\
  310 & 1 & \textcolor{teal}{\textbf{FAIR}} & \texttt{8.7.4} Amine Derivatizations & & \texttt{8.6.25} Hydrazone from active methylene \\
  311 & 1 & \textcolor{red}{\textbf{SUSPECT}} & \texttt{8.7.4} Amine Derivatizations & & \texttt{9.1.4} Halogenation of Heteroarenes \\
  312 & 1 & \textcolor{teal}{\textbf{FAIR}} & \texttt{8.7.6} Miscellaneous FGIs & & \texttt{5.1.3} Cleavage of Carbonyl PGs \\
  313 & 1 & \textcolor{teal}{\textbf{FAIR}} & \texttt{8.7.6} Miscellaneous FGIs & & \texttt{1.7.20} Hydrolysis of Aryl/Heteroaryl Halides \\
  314 & 1 & \textcolor{teal}{\textbf{FAIR}} & \texttt{8.7.6} Miscellaneous FGIs & & \texttt{1.7.1} Williamson Ether \\
  315 & 1 & \textcolor{teal}{\textbf{FAIR}} & \texttt{8.7.6} Miscellaneous FGIs & & \texttt{6.1.11} Reduction of nitrobenzenes \\
  316 & 1 & \textcolor{teal}{\textbf{FAIR}} & \texttt{8.7.6} Miscellaneous FGIs & & \texttt{8.5.1} Dehydration \\
  317 & 1 & \textcolor{teal}{\textbf{FAIR}} & \texttt{8.7.8} Decarboxylation & & \texttt{5.1.13} Decarboxylation and Decarbalkoxylation \\
  318 & 1 & \textcolor{red}{\textbf{SUSPECT}} & \texttt{8.7.8} Decarboxylation & & \texttt{6.9.6} Reduction of Carboxylic Acids \\
  319 & 1 & \textcolor{red}{\textbf{SUSPECT}} & \texttt{9.1.5} Halogenation of Arenes & & \texttt{1.7.2} O-Arylation of Alcohols and Phenols \\
  320 & 1 & \textcolor{red}{\textbf{SUSPECT}} & \texttt{9.1.5} Halogenation of Arenes & & \texttt{6.9.2} Reductive Cleavage and Decarboxylation \\
  321 & 1 & \textcolor{teal}{\textbf{FAIR}} & \texttt{9.1.6} Benzylic Halogenation & & \texttt{7.6.4} Halogenation of Benzylic C-H Bonds \\
\multicolumn{6}{l}{\small $^\dagger$ Sibling classes (same parent node).} \\
\end{longtable}

\clearpage
\section{Per-Template Coverage Distribution}
\label{sec:si_coverage}

\begin{figure}[htbp]
\centering
\includegraphics[width=\textwidth]{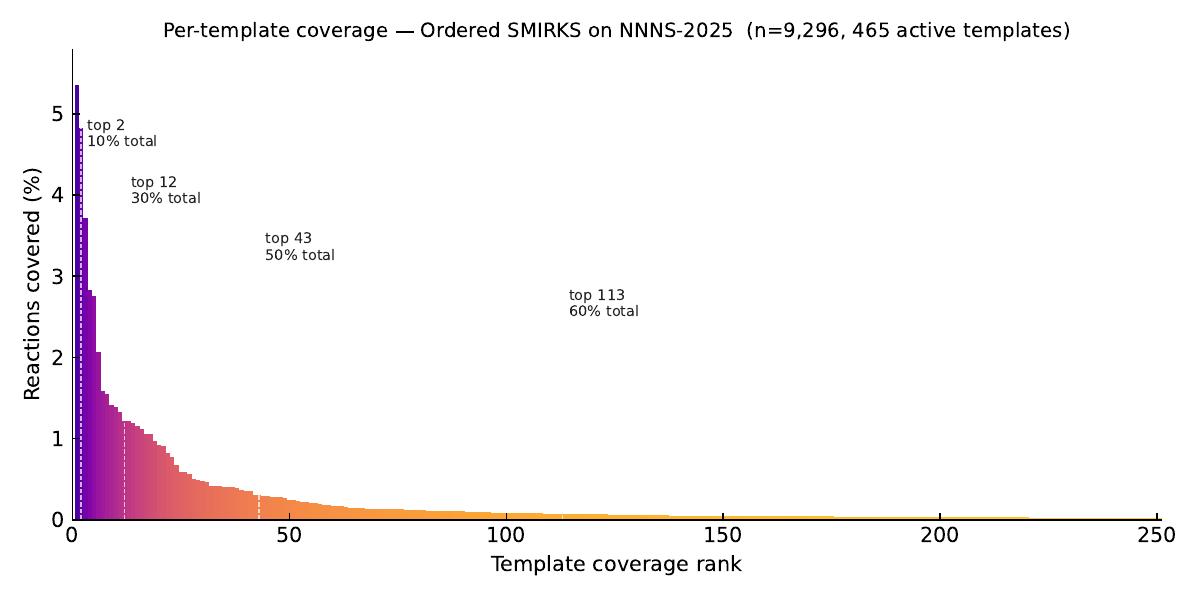}
\caption{Per-template coverage on the NNNS-2025 single-reaction-centre subset
  ($n=9{,}296$; 465 templates with at least one match).
  Each bar represents one Ordered SMIRKS template ranked by the number of reactions
  it classifies; bar colour encodes cumulative coverage reached at that rank
  (dark purple: low cumulative coverage; orange--yellow: approaching the
  $68.3\%$ ceiling).
  Milestone lines mark the top-2 ($10\%$), top-12 ($30\%$), top-43 ($50\%$),
  and top-113 ($60\%$) thresholds.
  The x-axis is truncated at rank 250; templates ranked 251--465 each contribute
  $\leq 0.03\%$ individually.}
\label{fig:si_coverage_curve}
\end{figure}

Figure~\ref{fig:si_coverage_curve} shows the individual contribution of each SMIRKS
template to coverage of the NNNS-2025 single-reaction-centre evaluation set
($n=9{,}296$), ranked in descending order of reactions classified.
The distribution follows a steep power law: the two most frequent templates
(Boc carbamate cleavage and Suzuki--Miyaura BPin coupling) together account for
$10\%$ of all reactions, 43 templates suffice to reach $50\%$, and 113 templates
are required for $60\%$.
Beyond rank $\sim$200 each additional template contributes fewer than $0.03\%$ of
reactions, illustrating the long-tail character of reaction-type frequency in
modern medicinal chemistry and motivating the LLM fallback described in the main text.

\end{document}